\documentclass{article}
\usepackage[T1]{fontenc}
\usepackage{microtype}
\usepackage{graphicx}
\usepackage{pgfplots}
\pgfplotsset{compat=1.18,theplot/.style={
  axis line style={black!70},tick style={black!70},
  tick label style={font=\scriptsize},label style={font=\small},
  title style={font=\small\bfseries},legend style={font=\scriptsize,draw=none,fill=none},
  grid=major,grid style={black!10},axis on top,
  every axis plot/.append style={line width=1pt},
  scaled ticks=false}}

\usepackage{placeins}
\usepackage{afterpage}
\usepackage{needspace}
\usepackage{booktabs}
\usepackage{array,multirow}
\usepackage{xcolor}
\usepackage{colortbl}
\definecolor{liftblue}{RGB}{53,104,168}
\definecolor{liftwash}{RGB}{239,246,255}
\definecolor{liftgreen}{RGB}{21,127,112}
\definecolor{liftred}{RGB}{181,77,69}
\usepackage{url}
\makeatletter
\g@addto@macro\UrlBreaks{\do\a\do\b\do\c\do\d\do\e\do\f\do\g\do\h\do\i\do\j\do\k\do\l\do\m\do\n\do\o\do\p\do\q\do\r\do\s\do\t\do\u\do\v\do\w\do\x\do\y\do\z\do\0\do\1\do\2\do\3\do\4\do\5\do\6\do\7\do\8\do\9}
\makeatother
\usepackage{hyperref}

\usepackage[preprint]{icml2027}
\usepackage{amsmath,amssymb,amsthm,mathtools}
\usepackage{algorithm}
\usepackage{algorithmic}
\usepackage[capitalize,noabbrev]{cleveref}
\DeclareRobustCommand{\panelref}[2]{\nameCref{#1}~\hyperref[#1]{\ref*{#1}#2}}

\newcommand{\method}{TransHarness}
\newcommand{\emptycell}{\textemdash}

\newcommand{\E}{\mathbb{E}}
\newcommand{\Prob}{\mathbb{P}}
\newcommand{\ind}{\mathbb{I}}
\icmltitlerunning{TransHarness: From Benchmark Overfitting to Transferable Harness Evolution}

\begin{document}
\twocolumn[
\icmltitle{TransHarness: From Benchmark Overfitting to Transferable Harness Evolution}
\icmlsetsymbol{equal}{\textdagger}
\icmlsetsymbol{corresponding}{*}
\begin{icmlauthorlist}
\icmlauthor{Tianze Xia}{hust,equal}
\icmlauthor{Zimo Wang}{hust,equal}
\icmlauthor{Heng Tong}{hust,equal}
\icmlauthor{Zhaoxiang Liu}{unicom}
\icmlauthor{Shiguo Lian}{unicom}
\icmlauthor{Fan Huang}{xmu}
\icmlauthor{Wenyu Liu}{hust}
\icmlauthor{Yunchao Wei}{bjtu}
\icmlauthor{Xinggang Wang}{hust,corresponding}
\end{icmlauthorlist}
\icmlaffiliation{hust}{Huazhong University of Science and Technology, Wuhan, China}
\icmlaffiliation{bjtu}{Beijing Jiaotong University
, Beijing, China}
\icmlaffiliation{xmu}{Xiamen University, Xiamen, China}
\icmlaffiliation{unicom}{Unicom Data Intelligence, China Unicom, Beijing, China}
\icmlcorrespondingauthor{Xinggang Wang}{xgwang@hust.edu.cn}
\icmlkeywords{agent harnesses, program evolution, language model agents, transfer}
\vskip 0.3in
]
\printAffiliationsAndNotice{\textsuperscript{\textdagger}Equal contribution. \textsuperscript{*}Corresponding author. Email to the first author: Tianze Xia \textless{}xiatianze@hust.edu.cn\textgreater{}. }
\flushbottom

\setcounter{dbltopnumber}{1}
\setcounter{topnumber}{2}

\begin{abstract}
Harness evolution offers a path to improving coding agents without updating model weights. However, existing methods can overfit their evolution benchmarks, producing gains that fail to carry over to new tasks. To address this challenge, we introduce \textbf{Trans}ferable \textbf{Harness} Evolution (\textbf{\method}), which evolves harness intervention policies from same-task success--failure contrasts. These policies couple execution-context triggers with diagnostic guidance, enabling task-dependent checks and corrections beyond the source executions. Cross-task replay and independent semantic selection retain policies that intervene selectively and ground corrections in current-task evidence rather than source-specific solutions. We further curate an independent evolution set for cross-distribution transfer and conduct a cross-benchmark evaluation covering multiple strong models and a range of lightweight and mature harnesses. Experiments show that, with only 300 source trajectories from outside the test distribution, \method\ further improves strong agents across distinct target benchmarks, demonstrating transferable harness evolution beyond benchmark overfitting. Specifically, \method\ achieves state-of-the-art scores of 76.4\% on DeepSWE v1.1 and 95.5\% on Terminal-Bench 2.1, surpassing the strongest baseline model, GPT-6 Astra, by 2.4 and \mbox{5.6 percentage points, respectively.}
\end{abstract}

\begin{figure*}[t] \centering \includegraphics[width=0.84\textwidth,trim=0 0 0 3bp,clip]{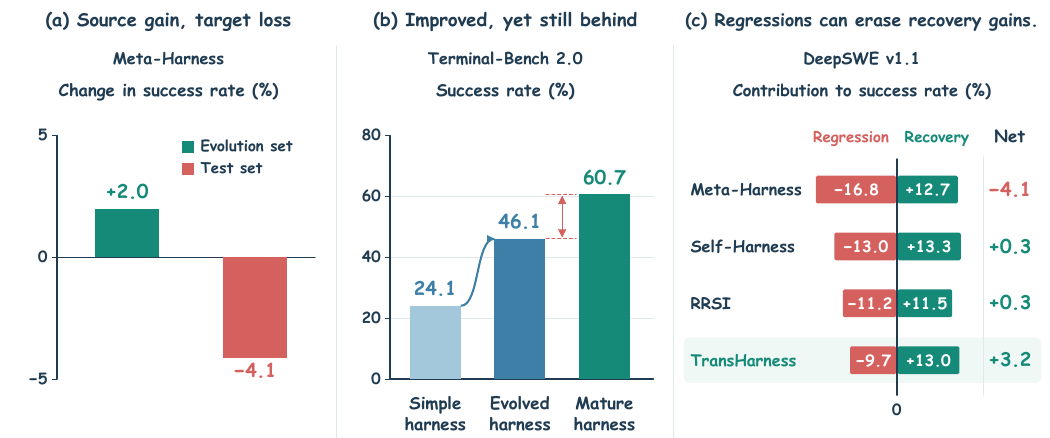} \caption{Challenges in transferable harness evolution. (a) Meta-Harness \citep{metaharness2026} improves on our independent evolution set but loses performance on DeepSWE v1.1 \citep{huang2026deepswev11}. (b) In HarnessOpt-Bench \citep{harnessoptbench2026}, an evolved simple Python harness still trails mature OpenCode under the same model and held-out tasks. The evolved score is reconstructed from the reported normalized gain. (c) Recovery gains and regression losses determine the net change in DeepSWE success (\cref{tab:methods}).} \label{fig:motivation} \end{figure*}

\section{Introduction}

Coding agents rely on harnesses to organize tool use, interpret feedback, and verify task outcomes \citep{sweagent2024,openhands2025}. Harness evolution improves these execution mechanisms using experience from past tasks, offering a path to stronger agents without updating model weights \citep{dgm2026,metaharness2026,modularrsi2026}.

However, gains on the tasks that drive evolution may fail to carry over to new tasks. SICA selects agent versions on a fixed task suite \citep{sica2025}, Meta-Harness searches for and evaluates coding harnesses on the same TerminalBench-2 tasks \citep{metaharness2026}, and Self-Harness selects candidates using feedback from held-out tasks within the benchmark \citep{selfharness2026}. Such feedback can favor adaptations to recurring benchmark characteristics, making benchmark overfitting difficult to distinguish from transferable improvement. DGM and ModularRSI test cross-benchmark transfer \citep{dgm2026,modularrsi2026}, yet sustaining gains under distribution shift remains challenging. Our Meta-Harness reproduction illustrates this gap. It improves on our independent evolution set, but its success rate falls on DeepSWE v1.1~(\panelref{fig:motivation}{a}).

To improve agents in practical settings, harness evolution must also benefit strong models running in mature harnesses. Existing methods can improve minimal starting systems, such as DGM's harness with basic execution and editing tools and Self-Harness's minimal DeepAgent \citep{dgm2026,selfharness2026}. How to improve mature systems remains an open question. In HarnessOpt-Bench, an evolved simple harness still trails mature OpenCode when both use the same model and evaluation tasks \citep{harnessoptbench2026} (\panelref{fig:motivation}{b}). As baseline success increases, fewer failures remain to recover and more successful executions are at risk of regression. A positive net gain on strong agents therefore requires a lower ratio of the regression rate to the recovery rate. If the ratio is too high, overall performance can decline even when evolution recovers some failures (\panelref{fig:motivation}{c}). To transfer gains to strong agents, interventions must uncover overlooked problems while preserving behavior that already works.

\begin{figure*}[t] \centering \includegraphics[width=0.94\textwidth]{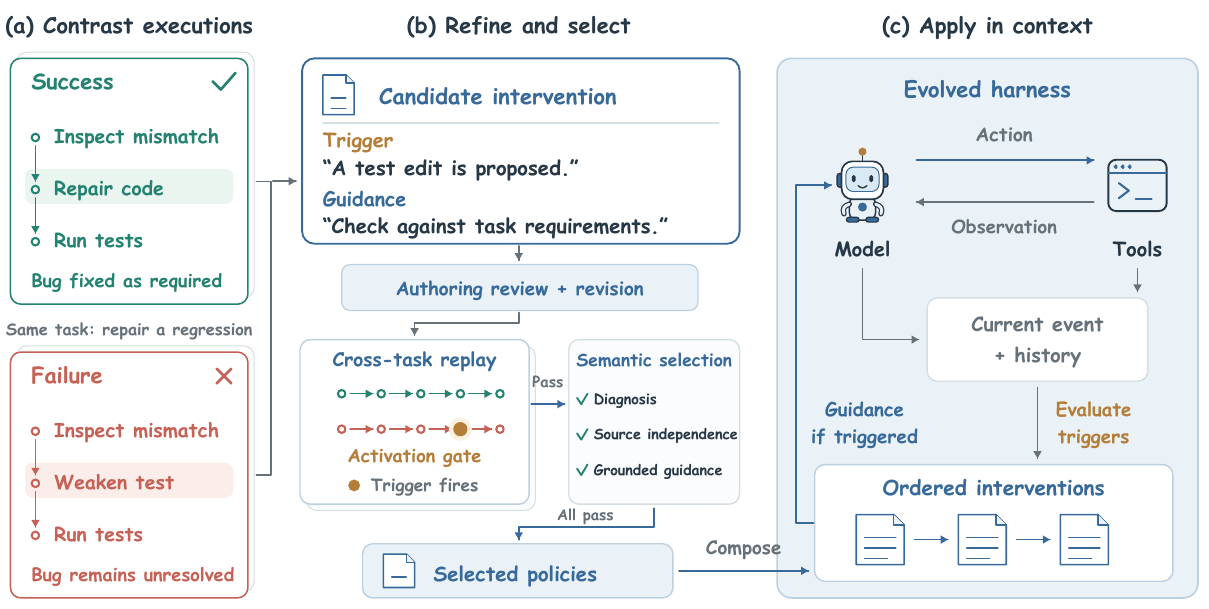} \caption{Schematic overview of transferable intervention evolution with \method. (a) Same-task contrasts motivate a candidate. (b) Authoring review refines the candidate. Candidates passing replay enter independent semantic selection of diagnosis, source independence, and evidence-grounded guidance; replay statistics are withheld. Gold nodes mark trigger activation. (c) Selected policies form a harness that responds to current events and history; \mbox{arrows indicate runtime information flow.}} \label{fig:overview} \end{figure*}

A correction can be specific to its source task even when the underlying execution failure recurs across task distributions. To address this challenge, we introduce \textbf{Trans}ferable \textbf{Harness} Evolution (\method), which evolves when and how a harness intervenes during execution. \method\ derives intervention policies from contrasts between successful and failed executions of the same task. Each policy pairs a trigger based on observable execution context with diagnostic guidance. The guidance directs agents to task requirements and evidence, allowing the same decision problem to receive different corrections across tasks. Cross-task replay and independent semantic selection together retain policies that intervene selectively and guide corrections based on evidence from the current task, making them usable beyond the \mbox{original source-task solution (\cref{fig:overview}).}

To test whether harness evolution transfers across task distributions, we construct an independent evolution set and characterize how it differs from the target benchmarks in code repositories, task types, and application domains. For each model and harness configuration, we run the baseline agent on the evolution set and use the resulting trajectories to generate and select intervention policies. We then evaluate the evolved systems on the target benchmarks, covering multiple strong models and both lightweight and mature harnesses. \method\ achieves state-of-the-art scores of 76.4\% on DeepSWE v1.1 and 95.5\% on Terminal-Bench 2.1 \citep{huang2026deepswev11,terminalbench2026v21}, exceeding GPT-6 Astra by 2.4 and \mbox{5.6 percentage points, respectively.}

Our main contributions are as follows.
\begin{itemize}

\item We find that gains on source tasks can turn into losses when transferred to a new benchmark. We derive the condition for harness evolution to yield a positive net gain on target tasks. At a fixed recovery rate, a higher baseline success rate allows a lower regression rate.
\item We propose \method, which evolves harness intervention policies from same-task success--failure contrasts. In relevant contexts, these policies guide agents to reassess decisions using current-task evidence. Cross-task replay tests how selectively they activate, and independent semantic selection checks whether diagnoses are supported and whether guidance \mbox{depends on source-task solutions.}
\setlength{\itemsep}{2pt}\setlength{\parskip}{0pt}
\item We construct an independent evolution set to evaluate transfer across task distributions. Experiments with multiple strong models in lightweight and mature harnesses demonstrate the transferable gains of \method, which achieves state-of-the-art performance on both DeepSWE v1.1 and Terminal-Bench 2.1.
\end{itemize}

\section{Related Work}
\paragraph{Learning and evolving agent harnesses.}
Harness evolution extends feedback-based optimization to executable agents \citep{agentoptimizer2024,aflow2025,adas2025}. Meta-Harness searches historical code, scores, and traces \citep{metaharness2026}; ModularRSI combines same-task contrasts, module-restricted updates, review, and task-aware composition \citep{modularrsi2026}. \method\ instead couples context-sensitive triggers with diagnostic guidance.

\noindent\textbf{Validation and transfer of harness improvements.}\quad
Self-Harness checks regressions \citep{selfharness2026}, VERSE co-evolves verification and optimization \citep{verse2026}, and RRSI regularizes selection \citep{rrsi2026}. Held-out evaluation separates search from testing \citep{rethinkingharnessevaluation2026,harnessoptbench2026,harnessdev2026,ding2026ade}. AHE verifies predicted edit effects from trajectory evidence \citep{ahe2026}; HarnessCompass constrains edits to task-agnostic mechanisms \citep{harnesscompass2026}; Meta$^2$ learns knowledge for task-local evolution \citep{meta22026}. \method\ instead freezes policies before deployment, using replay to test selectivity and semantic selection to examine source-specific assumptions. Further comparisons appear in \cref{app:related}.

\section{Transferable Harness Evolution}
\label{sec:method}
\label{sec:problem}
\label{sec:analysis}

\method\ evolves policies that recognize recurring decision problems and guide reassessment under new task requirements. Candidate refinement is followed by selection on source evidence (\cref{fig:overview}). \mbox{The selected policies are} composed into the final harness.

Given a fixed model $M$ and base harness $H_0$, we learn an ordered intervention set $B$ on source distribution $P$. The intervention set defines the evolved harness $H_B$, which uses the same execution model as the base harness for every new task.
The objective is to improve success on tasks drawn from an independent target distribution $Q$:
\begin{equation}
\Delta_Q(B)=
\E_{x\sim Q}\!\left[p_{H_B}(x)-p_{H_0}(x)\right],
\label{eq:target}
\end{equation}
Here, $p_H(x)$ is the probability that $H$ solves $x$. Construction and selection use source evidence alone; we freeze the resulting harness $H_B$ before evaluating it on target tasks.

\subsection{Evolving Intervention Policies}
\label{sec:content}

From a same-task success--failure pair $\rho=(\tau^+,\tau^-)$ under $H_0$, an author diagnoses divergent decisions and derives a reusable principle. Each pair starts from $H_0$; diagnoses without a reusable principle can be declined. \panelref{fig:overview}{a} contrasts repairing a bug with weakening a test. The resulting principle evaluates test changes against task requirements and the remaining defect, allowing task-dependent repairs.

\begin{figure*}[t]
\centering
\includegraphics[width=\textwidth,trim=5.7bp 0 5.7bp 0,clip]{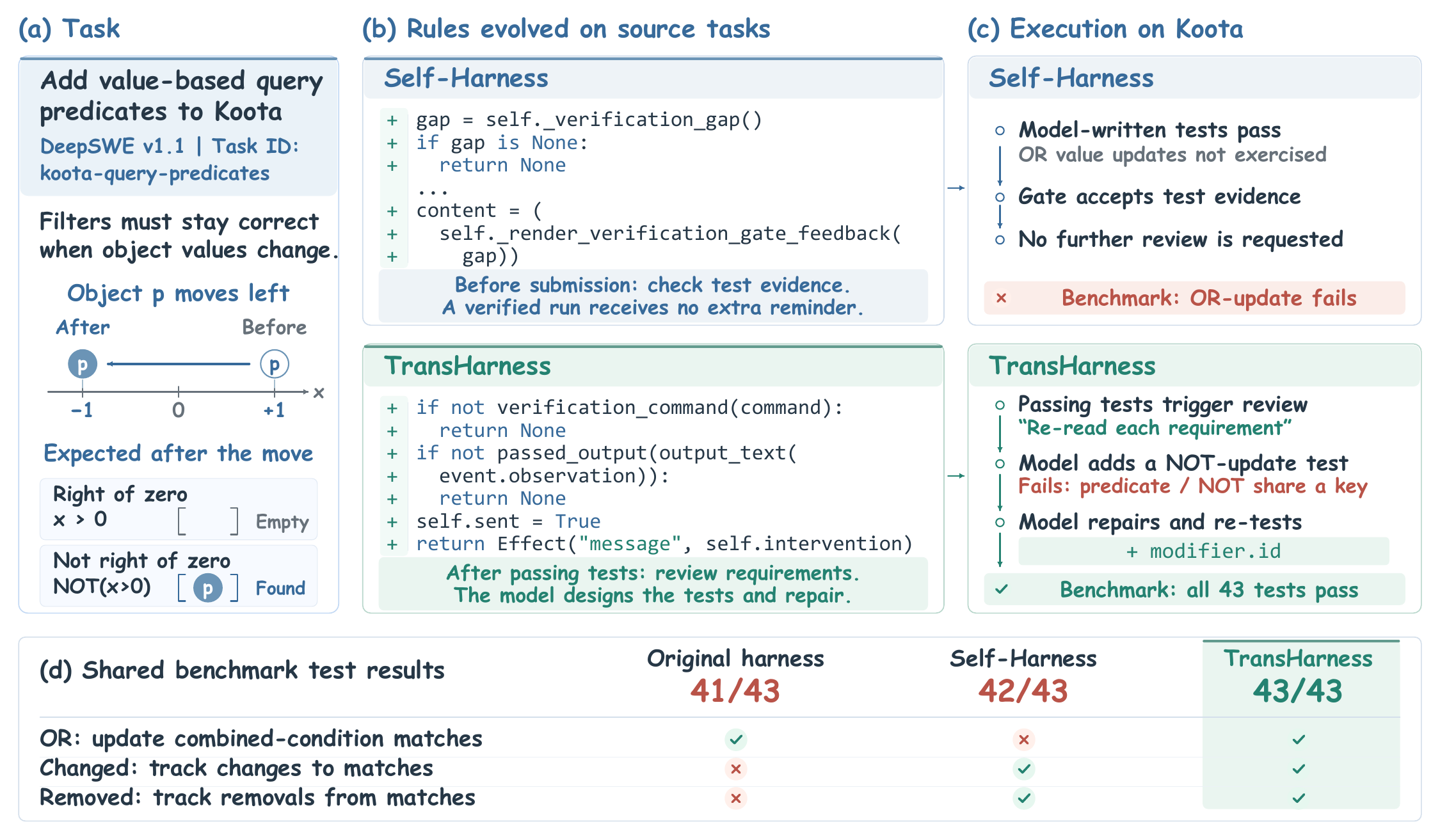}
\caption{Transfer of an evolved decision principle to a Koota task. (a) Target requirements. (b) Source-evolved code from the adapted Self-Harness candidate \citep{selfharness2026} and \method. (c) Self-Harness checks verification evidence before submission; \method\ prompts a requirement audit after passing checks, leading to a new NOT-update test and cache-key repair. (d) Native evaluation of independent executions on the same 43 task-specific tests; all executions pass the same 172 tests for regressions.}
\label{fig:generalization}
\end{figure*}

Each policy $u=(s,g,c)$ specifies a harness location $s$, executable trigger $g$, and diagnostic guidance $c$. When $g(h_t,e_t)=1$ for history $h_t$ and event $e_t$, the harness supplies $c$ through $s$ to direct reassessment toward relevant requirements and evidence. Locations cover model calls, tools, context updates, and completion (\cref{app:slots}). Triggers use observable execution context; the policy transfers a diagnostic procedure whose resolution remains responsive to current requirements (\cref{fig:generalization}).
\begin{algorithm}[!t] \caption{\method\ self-evolution} \label{alg:construct} \begin{algorithmic}[1] \REQUIRE Base harness $H_0$, source pairs $\Pi$, histories $\mathcal R$ \STATE $I\leftarrow[\,]$ \FOR{each $\rho_i\in\Pi$ in source order} \STATE $z_i\leftarrow\mathrm{Author}(\rho_i,H_0)$ \STATE $(z_i,d_i)\leftarrow\mathrm{Review}(\rho_i,z_i)$ \STATE \textbf{if} $d_i=\mathrm{rework}$ \textbf{then} $z_i\leftarrow\mathrm{Rework}(\rho_i,z_i)$ \STATE $(h_i,\mathbf n_i)\leftarrow\mathrm{CheckReplay}(z_i,\mathcal R)$ \IF{$h_i=1\;\land\;u(z_i)\neq\varnothing$} \STATE $I\leftarrow I\mathbin{\|}[i]$ \ENDIF \ENDFOR \STATE $I_G\leftarrow[i\in I:G(\mathbf n_i)=1]$ \FOR{each $i\in I_G$} \STATE $(v_{\mathrm d i},v_{\mathrm s i},v_{\mathrm a i})\leftarrow\mathrm{SelectSemantic}(\rho_i,z_i)$ \ENDFOR \STATE $J\leftarrow[i\in I_G:v_{\mathrm d i}v_{\mathrm s i}v_{\mathrm a i}=1]$ \STATE \textbf{return} $\mathrm{Compose}([u(z_j)]_{j\in J},H_0)$ \end{algorithmic} \end{algorithm}

\subsection{Refining Intervention Scope and Semantics}
\label{sec:replay}

Two properties determine whether a source-derived intervention remains useful: which executions it affects, and what its guidance induces in those executions. Consider $H$ and $H+u$ coupled through common randomness, with identical histories and resource-limit effects until $u$ first affects execution. Let $A_u$ denote this event. For distribution $D\in\{P,Q\}$ with $\Prob_D(A_u)>0$, define exposure $a_D=\Prob_D(A_u)$ and conditional value $v_D=\E_D[Y_{H+u}-Y_H\mid A_u]$. Only executions in $A_u$ can \mbox{contribute to the difference, giving}
\begin{equation}
\Delta_D(u\mid H)=a_Dv_D.
\label{eq:scope}
\end{equation}
The source--target change therefore satisfies
\begin{equation}
\begin{multlined}
\Delta_Q(u\mid H)-\Delta_P(u\mid H)\\
=(a_Q-a_P)v_P+a_Q(v_Q-v_P).
\end{multlined}
\label{eq:transfer-gap}
\end{equation}
Transfer can fail through either factor: an unchanged activation rate can accompany inappropriate corrections, while overly narrow activation misses recoverable failures. Both scope and guidance must generalize beyond the conditions encountered in the source task. The decomposition motivates two complementary checks: replay examines where a policy intervenes, while semantic selection examines whether its guidance remains valid under new requirements.

The consequences are sharper for strong baselines. Let $(Y_0,Y_B)$ be paired binary outcomes on $Q$, with base success $p_0=\Prob(Y_0=1)$, rescue rate $r=\Prob(Y_B=1\mid Y_0=0)$, and regression rate $q=\Prob(Y_B=0\mid Y_0=1)$. Then
\begin{equation}
\Delta=(1-p_0)r-p_0q.
\label{eq:balance}
\end{equation}
Even perfect recovery is limited by the failures remaining under the base harness. For $0<p_0<1$, \cref{eq:balance} gives this ceiling and requires recovery gains to outweigh losses on originally successful executions:
\begin{equation}
\Delta\leq1-p_0,\qquad
\Delta>0\iff q<\frac{1-p_0}{p_0}r.
\label{eq:constraints}
\end{equation}
Higher baseline success leaves fewer failures to recover and lowers the allowable regression-to-rescue ratio. A fixed increase in $q$ then costs more success points. Useful interventions reach correctable decisions while preserving successful behavior. \Cref{app:proofs} gives the derivations and states the assumptions on coupled executions.

Authoring review refines each candidate against its motivating pair (\panelref{fig:overview}{b}). The reviewer revises guidance wording or requests one author rework for changes to the diagnosis or trigger. Executable checks validate the implementation and record activation and control-flow effects on both successful and failed source histories.

For example, a completion intervention initially required running existing tests and ``fixing any failures.'' Review identified an appropriate execution that left an unrelated failure unresolved. The revised guidance repairs failures caused by the current change and reports unrelated ones, preserving verification without an unconditional repair obligation. Independent selection then assesses whether the completed candidate should enter the final composition.

\subsection{Transfer-Oriented Policy Selection}
\label{sec:selection}

Selection screens event-triggered activation and assesses corrective obligations. For each policy $u$, let $\mathbf n(u)=(n_{-},n_{+},n_{\pm},N)$ summarize replay on $N>0$ trajectory pairs from other source tasks. The first three counts indicate activation on failed executions only, successful executions only, or both, excluding session-start guidance. \mbox{The replay gate is}
\begin{equation}
G(\mathbf n(u))=\ind\!\left[\frac{n_{\pm}}{N}<0.9\right]
      \ind[n_{-}\geq n_{+}].
\label{eq:selection-gate}
\end{equation}
It rejects nearly indiscriminate event-triggered activation and policies that more often single out successful executions. Surviving policies, including session-start guidance, then undergo independent semantic assessment of the frozen candidate and its motivating pair, without replay statistics. Its three verdicts $v_{\mathrm d},v_{\mathrm s},v_{\mathrm a}$ require a supported diagnosis, independence from the source solution, and corrective actions grounded in failed checks or explicit \mbox{task requirements. Selection retains}
\begin{equation}
\mathcal U^*=\{u\in\mathcal U:G(\mathbf n(u))v_{\mathrm d}(u)v_{\mathrm s}(u)v_{\mathrm a}(u)=1\}.
\label{eq:selected-bundle}
\end{equation}
The two stages assess event-triggered selectivity and task-grounded guidance: a credible diagnosis can trigger indiscriminately, while a selective trigger can carry an unjustified correction. The selector accepts or rejects without editing. Selected policies are composed in source order within each interface (\panelref{fig:overview}{c}) and frozen before target evaluation. Because interventions alter later contexts, we evaluate the complete harness (\cref{eq:composition}). Algorithm~\ref{alg:construct} uses proposal $z_i$, checker acceptance $h_i$, and replay counts $\mathbf n_i=\mathbf n(u(z_i))$; \cref{app:modes} details the judgments.

\begin{table*}[!t]
\centering
\begin{minipage}{\textwidth}
\caption{Results on DeepSWE v1.1 (113 tasks) and Terminal-Bench 2.1 (89 tasks). Harnesses separated by a slash follow benchmark order. Public entries retain their reported settings (\cref{tab:public-configs}).}
\label{tab:main}
\begingroup
\fontsize{9.9}{11.88}\selectfont
\setlength{\tabcolsep}{4pt}
\renewcommand{\arraystretch}{1.017}
\begin{tabular}{@{\hspace{8pt}}>{\centering\arraybackslash}p{\dimexpr.34\linewidth-14pt\relax}>{\centering\arraybackslash}p{\dimexpr.24\linewidth-10pt\relax}>{\centering\arraybackslash}p{\dimexpr.21\linewidth-8pt\relax}>{\centering\arraybackslash}p{\dimexpr.21\linewidth-8pt\relax}@{\hspace{8pt}}}
\toprule[0.8pt]
\raisebox{-0.5\baselineskip}[0pt][0pt]{Model} & \raisebox{-0.5\baselineskip}[0pt][0pt]{Harness} & \multicolumn{2}{c}{Score (\%)} \\
\cmidrule(lr){3-4}
 & & DeepSWE v1.1 & Terminal-Bench 2.1 \\
\midrule
Gemini 3.1 Pro & \mbox{Mini-SWE\hspace{2pt}/\hspace{2pt}Terminus 2} & 12.0 & 73.8 \\
Muse Spark 1.1 & \mbox{Mini-SWE\hspace{2pt}/\hspace{2pt}Terminus 2} & 53.0 & 77.9 \\
Grok 4.5 & \mbox{Mini-SWE\hspace{2pt}/\hspace{2pt}Terminus 2} & 54.0 & 81.6 \\
Claude Sonnet 5 & \mbox{Mini-SWE\hspace{2pt}/\hspace{2pt}Terminus 2} & 54.0 & 80.5 \\
Claude Opus 4.8 & \mbox{Mini-SWE\hspace{2pt}/\hspace{2pt}Terminus 2} & 59.0 & 84.6 \\
GPT-5.5 & \mbox{Mini-SWE\hspace{2pt}/\hspace{2pt}Terminus 2} & 67.0 & 84.3 \\
Claude Fable 5 & \mbox{Mini-SWE\hspace{2pt}/\hspace{2pt}Terminus 2} & 70.0 & 84.6 \\
GPT-5.6 Sol & \mbox{Mini-SWE\hspace{2pt}/\hspace{2pt}Terminus 2} & 73.0 & 89.5 \\
Claude Opus 5 & \mbox{Mini-SWE\hspace{2pt}/\hspace{2pt}Terminus 2} & 74.0 & 89.1 \\
GPT-6 Astra & \mbox{Mini-SWE\hspace{2pt}/\hspace{2pt}Terminus 2} & 74.0 & 89.9 \\
\midrule[0.4pt]
\raisebox{\dimexpr-1.5\baselineskip-2.8pt\relax}[0pt][0pt]{Qwen3.8-Flash-Next} & \mbox{Mini-SWE\hspace{2pt}/\hspace{2pt}Terminus 2} & 65.5 & 85.4 \\
 & \textbf{+TransHarness} & \textbf{69.0\,\textcolor{liftgreen}{(+3.5)}} & \textbf{86.5\,\textcolor{liftgreen}{(+1.1)}} \\
\addlinespace[3pt]
 & Qwen Code & 55.8 & 86.5 \\
 & \textbf{+TransHarness} & \textbf{63.7\,\textcolor{liftgreen}{(+7.9)}} & \textbf{89.9\,\textcolor{liftgreen}{(+3.4)}} \\
\cmidrule(lr){1-4}
\raisebox{\dimexpr-1.5\baselineskip-2.8pt\relax}[0pt][0pt]{GLM-5.3 Flash} & \mbox{Mini-SWE\hspace{2pt}/\hspace{2pt}Terminus 2} & 63.4 & 77.5 \\
 & \textbf{+TransHarness} & \textbf{67.3\,\textcolor{liftgreen}{(+3.9)}} & \textbf{82.0\,\textcolor{liftgreen}{(+4.5)}} \\
\addlinespace[3pt]
 & ZCode & 60.2 & 82.0 \\
 & \textbf{+TransHarness} & \textbf{61.9\,\textcolor{liftgreen}{(+1.7)}} & \textbf{83.1\,\textcolor{liftgreen}{(+1.1)}} \\
\cmidrule(lr){1-4}
\raisebox{\dimexpr-1.5\baselineskip-2.8pt\relax}[0pt][0pt]{DeepSeek V4.1 Flash (1017)} & \mbox{Mini-SWE\hspace{2pt}/\hspace{2pt}Terminus 2} & 73.2 & 84.3 \\
 & \textbf{+TransHarness} & \textbf{76.4\,\textcolor{liftgreen}{(+3.2)}} & \textbf{86.5\,\textcolor{liftgreen}{(+2.2)}} \\
\addlinespace[3pt]
 & DeepSeek Harness & 70.8 & 89.9 \\
 & \textbf{+TransHarness} & \textbf{75.2\,\textcolor{liftgreen}{(+4.4)}} & \textbf{95.5\,\textcolor{liftgreen}{(+5.6)}} \\
\bottomrule[0.8pt]
\end{tabular}
\endgroup
\end{minipage}
\par\medskip
\begin{minipage}[t]{\columnwidth}
\phantomsection
\makeatletter\let\stackedtableanchor\@currentHref\makeatother
\centering
\caption{Transfer to DeepSWE v1.1 after evolving Mini-SWE with a shared source archive and executor. Parentheses show changes from the base harness.}
\makeatletter\let\@currentHref\stackedtableanchor\makeatother
\label{tab:methods}
\footnotesize
\setlength{\tabcolsep}{1.5pt}
\renewcommand{\arraystretch}{1.15}
\begin{tabular}{@{}>{\centering\arraybackslash}m{\dimexpr.40\linewidth-\tabcolsep\relax}>{\centering\arraybackslash}m{\dimexpr.28\linewidth-2\tabcolsep\relax}>{\centering\arraybackslash}m{\dimexpr.16\linewidth-2\tabcolsep\relax}>{\centering\arraybackslash}m{\dimexpr.16\linewidth-\tabcolsep\relax}@{}}
\toprule
Method & Score $\uparrow$ & Rec. $\uparrow$ & Reg. $\uparrow$ \\
\midrule
Base harness & 73.2 & {---} & {---} \\
Meta-Harness & \makebox[1.9em][r]{69.0}\,\makebox[3.1em][l]{\textcolor{liftred}{(\textminus{}4.1)}} & +12.7 & \textminus{}16.8 \\
Self-Harness & \makebox[1.9em][r]{73.5}\,\makebox[3.1em][l]{\textcolor{liftgreen}{(+0.3)}} & \textbf{+13.3} & \textminus{}13.0 \\
RRSI & \makebox[1.9em][r]{73.5}\,\makebox[3.1em][l]{\textcolor{liftgreen}{(+0.3)}} & +11.5 & \textminus{}11.2 \\
TransHarness (Ours) & \makebox[1.9em][r]{\textbf{76.4}}\,\makebox[3.1em][l]{\textcolor{liftgreen}{(+3.2)}} & +13.0 & \textbf{\textminus{}9.7} \\
\bottomrule
\end{tabular}

\end{minipage}\hfill
\begin{minipage}[t]{\columnwidth}
\phantomsection
\makeatletter\let\stackedtableanchor\@currentHref\makeatother
\centering
\caption{Transfer to DeepSWE v1.1 under source budgets (\cref{app:efficiency}). Each task supplies three trajectories; parentheses show changes from the base harness.}
\makeatletter\let\@currentHref\stackedtableanchor\makeatother
\label{tab:efficiency}
\footnotesize
\setlength{\tabcolsep}{1.5pt}
\renewcommand{\arraystretch}{1.15}
\begin{tabular}{@{}>{\centering\arraybackslash}m{\dimexpr.40\linewidth-\tabcolsep\relax}>{\centering\arraybackslash}m{\dimexpr.28\linewidth-2\tabcolsep\relax}>{\centering\arraybackslash}m{\dimexpr.16\linewidth-2\tabcolsep\relax}>{\centering\arraybackslash}m{\dimexpr.16\linewidth-\tabcolsep\relax}@{}}
\toprule
Source tasks & Score $\uparrow$ & Rec. $\uparrow$ & Reg. $\uparrow$ \\
\midrule
0 & 73.2 & {---} & {---} \\
25 & \makebox[1.9em][r]{69.9}\,\makebox[3.1em][l]{\textcolor{liftred}{(\textminus{}3.2)}} & +12.1 & \textminus{}15.3 \\
50 & \makebox[1.9em][r]{70.8}\,\makebox[3.1em][l]{\textcolor{liftred}{(\textminus{}2.4)}} & \textbf{+14.7} & \textminus{}17.1 \\
75 & \makebox[1.9em][r]{74.3}\,\makebox[3.1em][l]{\textcolor{liftgreen}{(+1.2)}} & +14.2 & \textminus{}13.0 \\
100 & \makebox[1.9em][r]{\textbf{76.4}}\,\makebox[3.1em][l]{\textcolor{liftgreen}{(+3.2)}} & +13.0 & \textbf{\textminus{}9.7} \\
\bottomrule
\end{tabular}

\end{minipage}
\end{table*}

\subsection{Constructing Evolution Evidence}
\label{sec:source-construction}

The evolution set must provide informative failures and varied contexts for examining whether a diagnosis extends beyond its motivating task. We construct it from SWE-bench Pro \citep{swebenchpro2026}, combining difficulty-ranked selection with repository coverage. For repository pool $\mathcal P_j$, public multi-model solve rate $r_{\mathrm{pub}}(x)$, and fixed language offset $\beta_{\ell(x)}$, the ranking score and coverage constraints are
\begin{equation}
\begin{gathered}
\kappa(x)=r_{\mathrm{pub}}(x)+\beta_{\ell(x)},\\
|\mathcal S|=n,\qquad |\mathcal S\cap\mathcal P_j|\geq m\quad\forall j.
\end{gathered}
\label{eq:source-selection}
\end{equation}
Low $\kappa$ prioritizes low historical solve rates, adjusted by language preference. We select within repository difficulty bands, fill in ascending $\kappa$ order, and enforce coverage. With $n=100$ and $m=4$, the set spans eleven repositories; three executions per task provide 300 trajectories for outcome contrasts and cross-task refinement.

The set retains SWE-bench Pro's executable environments and verifiers \citep{swebenchpro2026}. We check execution readiness for all 100 tasks and inspect instruction--test alignment in a language-stratified sample covering the four source-language groups. Selection parameters and quality checks appear in \cref{app:source-protocol,app:source-quality}.

\begin{figure*}[t]
\centering
\includegraphics[width=\textwidth,height=200.17pt,keepaspectratio]{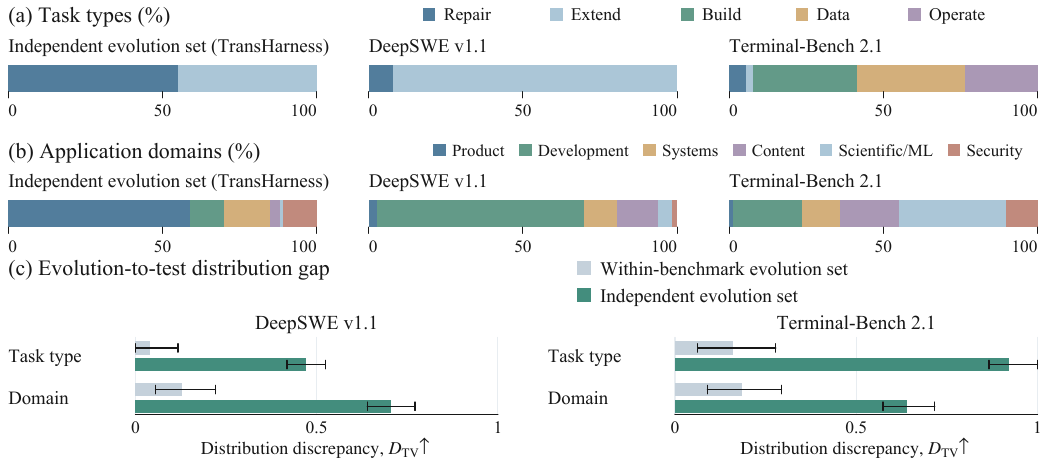}
\caption{Source--target distribution shift. (a,b) Task types and application domains of our independent evolution set and both target benchmarks. (c) Total variation $D_{\mathrm{TV}}$ (\cref{eq:task-shift}) compares the same target subset with a benchmark-internal or our independent evolution set. Bars average 10,000 random 2:1 partitions; whiskers show the 5th--95th percentiles. Larger distances indicate greater shift. Annotation and partition details appear in \cref{app:distribution}.}
\label{fig:source-distribution}
\end{figure*}

After fixing $\mathcal S$, we characterize its relationship to the target benchmarks through task type (the requested operation) and application domain (its intended use). Model-assisted annotation assigns these labels under a common codebook. For either view $f$, category proportions $p_D^f(z)$ in task set $D$ define the total variation (TV) distance:
\begin{equation}
D_{\mathrm{TV}}^f(\mathcal S,\mathcal T)=\frac12\sum_z\left|p_{\mathcal S}^f(z)-p_{\mathcal T}^f(z)\right|.
\label{eq:task-shift}
\end{equation}
TV is zero for identical proportions and one for disjoint categories (\cref{fig:source-distribution,sec:source-target}).

\section{Experiments}
\label{sec:experiments}

We first measure how well \method\ transfers across benchmarks, models, and harnesses. We then compare evolution methods using the same model, source trajectories, and initial harness, and use ablations to identify which design choices support transfer.

\subsection{Experimental Setup}

\noindent\textbf{Evolution data and target benchmarks.}\enspace
For each configuration, we run the base agent three times on each task in our independent 100-task evolution set (\cref{sec:source-construction}). \method\ constructs and selects interventions from the resulting 300 trajectories. We freeze each evolved harness before testing on DeepSWE v1.1 (113 tasks) \citep{deepswe2026,huang2026deepswev11} or on Terminal-Bench 2.1 (89 tasks) \citep{terminalbench2026,terminalbench2026v21}.

\noindent\textbf{Models and harnesses.}\enspace
We evaluate DeepSeek V4.1 Flash (1017) \citep{deepseek2026flash1017}, GLM-5.3 Flash \citep{zai2026glm53flash}, and Qwen3.8-Flash-Next \citep{qwen2026flashnext}. Our lightweight harnesses are Mini-SWE on DeepSWE and Terminus 2 on Terminal-Bench \citep{miniswe2026,terminus2026}. Native harnesses include DeepSeek Harness, ZCode, and Qwen Code \citep{deepseekharness2026,zcode2026,qwencode2026}. We fix the model settings, tools, environment, and execution limits when comparing the base and evolved harnesses.

\noindent\textbf{Comparison settings.}\enspace
Table~\ref{tab:main} compares base and evolved configurations on both benchmarks alongside public scores \citep{datacurve2026leaderboard,artificialanalysis2026tb21}. On DeepSWE v1.1, we compare \method, Meta-Harness, Self-Harness, and RRSI using the same DeepSeek V4.1 Flash (1017) model, initial Mini-SWE harness, and source trajectories from this configuration, with method-specific search budgets \citep{metaharness2026,selfharness2026,rrsi2026}. For RRSI, we report the target score of the candidate with the highest source-task success rate. Candidate selection is detailed in \cref{app:compared-methods}.

\noindent\textbf{Score reporting.}\enspace
All our evaluations report means over three independent runs ($N=3$); public scores come from official reports or public leaderboards. For \cref{tab:methods,tab:efficiency,tab:mechanism}, we compute scores, recovery and regression contributions, and net gains from the underlying execution outcomes at full precision, then round each reported quantity independently to one decimal place. Subtracting displayed scores or summing displayed recovery and regression contributions may therefore differ from the reported net gain \mbox{by 0.1 percentage points.}

\subsection{Main Results}
\label{sec:source-target}

\noindent\textbf{Transfer across benchmarks and harnesses.}\enspace
\Cref{tab:main} shows that \method\ improves DeepSeek, Qwen, and GLM on both benchmarks, with gains across lightweight and native harnesses. On DeepSWE v1.1, improvements hold for Mini-SWE and all three native harnesses, with Qwen Code gaining the most at 7.9 points. On Terminal-Bench 2.1, all three models likewise improve with either Terminus 2 or their native harnesses. DeepSeek with Mini-SWE and DeepSeek Harness achieves the best scores on the respective benchmarks, 76.4\% and 95.5\%, surpassing GPT-6 Astra by 2.4 and 5.6 points. For these two configurations, paired gains over their respective base harnesses have 95\% CIs of $[-2.4,\,8.8]$ and $[-1.1,\,12.4]$ points~(\cref{tab:requirement-transfer}).

\noindent\textbf{Transfer under distribution shift.}\enspace
Our independent evolution set differs from the target benchmarks in both repository coverage and task distribution. It shares none of DeepSWE's 91 repositories and emphasizes repair and extension in product workflows. DeepSWE centers on software development, while Terminal-Bench spans build, data production, and system configuration \citep{terminalbench2026v21} (\panelref{fig:source-distribution}{a,b}). With the target subset fixed, replacing a benchmark-internal evolution set \citep{selfharness2026} with our independent set increases mean domain-distribution distance from 0.13 to 0.70 on DeepSWE and from 0.18 to 0.64 on Terminal-Bench. The gap persists at matched evolution-set sizes (\panelref{fig:source-distribution}{c}). We further examine transfer to unseen task compositions. \method\ improves success on compositional OOD tasks in all four DeepSeek configurations, with Mini-SWE and Terminus 2 gaining 14.6 and 4.1 points, respectively (\cref{tab:requirement-transfer}). These gains show that evolved policies can apply to requirement combinations absent \mbox{from the source tasks.}

\subsection{Comparison with Harness Evolution Methods}
Starting from the same Mini-SWE base, the methods obtain different cross-benchmark gains from source-task evolution. Meta-Harness \citep{metaharness2026} gains 2.0 points on the evolution set but loses 4.1 on DeepSWE (\panelref{fig:motivation}{a,c}). Self-Harness and RRSI recover failures in 13.3\% and 11.5\% of all paired executions, respectively, but regress in 13.0\% and 11.2\%, leaving each with a net gain of only 0.3 points (\cref{tab:methods}). In comparison, \method\ recovers 13.0\%, close to Self-Harness, while reducing regressions to 9.7\%, the lowest among the compared evolution methods. By retaining recovery gains while disrupting fewer successful executions, \method\ reaches 76.4\% on DeepSWE, exceeding the base harness and strongest comparator by 3.2 and 2.9 points, respectively. \Cref{app:paired-outcomes} reports paired confidence intervals and scores for all three evaluation runs.

\subsection{Ablation Analysis}
\vspace{-3.24pt}

\noindent\textbf{Intervention timing.}\quad
We first examine whether guidance depends on the execution context in which it is delivered (\cref{tab:mechanism}). Runtime variants share interfaces, policy order, and activation caps, with similar guidance lengths. Delivering the same guidance at the first eligible event rather than the learned trigger reduces success to 72.6\%. Recovery remains close to the full method, 12.7\% versus 13.0\%, but regressions rise from 9.7\% to 13.3\%. Thus, learned timing preserves much of the guidance's corrective benefit while reducing disruption to \mbox{executions that would otherwise succeed.}

\noindent\textbf{Diagnostic guidance.}\quad
Retaining the learned trigger and diagnostic topic, we replace evidence-based reassessment with a prescribed correction. Success falls to 65.5\%, recovery drops to 8.8\%, and regressions rise to 16.5\%. Combining direct correction with early activation yields the highest regression rate, 17.7\%. Together, these controls explain how intervention design shapes the balance in \cref{eq:balance}. The trigger identifies a relevant execution context, while diagnostic guidance directs the agent to resolve the issue using current requirements and evidence. Fixing the correction in advance weakens both failure recovery and \mbox{preservation of successful behavior.}

\begin{table}[t] \begingroup \phantomsection \makeatletter\let\stackedtableanchor\@currentHref\makeatother \centering
\caption{Ablations on DeepSWE v1.1. The no-review variant uses a separately constructed intervention set. Parentheses show score changes relative to \method.}
\makeatletter\let\@currentHref\stackedtableanchor\makeatother
\label{tab:mechanism}
\label{tab:construction}
\small
\setlength{\tabcolsep}{1.5pt}
\renewcommand{\arraystretch}{1.54}
\begin{tabular}{@{}>{\raggedright\arraybackslash}m{\dimexpr.50\linewidth-\tabcolsep\relax}>{\centering\arraybackslash}m{\dimexpr.24\linewidth-2\tabcolsep\relax}>{\centering\arraybackslash}m{\dimexpr.13\linewidth-2\tabcolsep\relax}>{\centering\arraybackslash}m{\dimexpr.13\linewidth-\tabcolsep\relax}@{}}
\toprule
Method & Score $\uparrow$ & Rec. $\uparrow$ & Reg. $\uparrow$ \\
\midrule
\textbf{TransHarness (Ours)} & \textbf{76.4} & \textbf{+13.0} & \textbf{\textminus{}9.7} \\
\hspace*{6pt}w/o learned triggers & \makebox[1.9em][r]{72.6}\,\makebox[3.1em][l]{\textcolor{liftred}{(\textminus{}3.8)}} & +12.7 & \textminus{}13.3 \\
\hspace*{6pt}w/o reassessment & \makebox[1.9em][r]{65.5}\,\makebox[3.1em][l]{\textcolor{liftred}{(\textminus{}10.9)}} & +8.8 & \textminus{}16.5 \\
\hspace*{6pt}w/o triggers \& reassessment & \makebox[1.9em][r]{66.4}\,\makebox[3.1em][l]{\textcolor{liftred}{(\textminus{}10.0)}} & +10.9 & \textminus{}17.7 \\
\hspace*{6pt}w/o construction review & \makebox[1.9em][r]{67.3}\,\makebox[3.1em][l]{\textcolor{liftred}{(\textminus{}9.1)}} & +9.4 & \textminus{}15.3 \\
\bottomrule
\end{tabular}
 \endgroup \end{table}

\noindent\textbf{Construction review.}\quad
Construction review examines the diagnostic basis and scope of candidate interventions before deployment. The separately constructed harness with fourteen unreviewed interventions reaches 67.3\%, whereas the reviewed seven-intervention harness reaches 76.4\%, with more recoveries and fewer regressions (\cref{tab:mechanism}). The example in \cref{sec:replay} illustrates the role of this refinement. Review preserves the need for verification while restricting repair to failures caused by the current code change, avoiding guidance that would turn unrelated \mbox{failures into unnecessary edits.}

\subsection{Additional Analyses}

\noindent\textbf{Policy transfer on Koota.}\quad
The Koota case shows how a source-evolved policy adapts to a new task (\cref{fig:generalization}). After local tests pass, the policy prompts the agent to reassess requirement coverage, revealing untested NOT-update behavior. The agent then adds a test and repairs the cache key. The base harness passes 41 of 43 task-specific tests, while Self-Harness, which checks verification evidence before completion \citep{selfharness2026}, passes 42. \method\ passes all 43, and all three harnesses pass the 172 regression tests. Here, the source policy provides a reason to question incomplete verification, while the target task supplies the requirement and evidence needed to derive the repair.

\begin{figure}[t]
\centering
\includegraphics[width=\columnwidth,trim=0 0 0 6bp,clip]{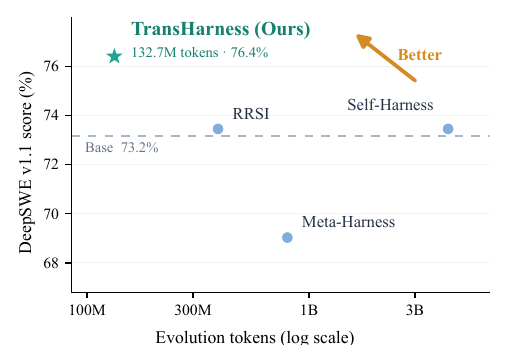}
\caption{DeepSWE v1.1 success versus evolution tokens (\cref{tab:methods}). Cost includes candidate construction and source validation; it excludes shared archive collection and evaluation on the target benchmark (\cref{app:evolution-tokens}).}
\label{fig:evolution-efficiency}
\end{figure}

\noindent\textbf{Source budget.}\quad
We compare harnesses constructed from nested subsets of the independent evolution set. Harnesses built from 25 and 50 tasks fall below the base, while those built from 75 and 100 tasks gain 1.2 and 3.2 points (\cref{tab:efficiency}). The 50-task harness recovers more failures than the 100-task harness, yet incurs a larger regression loss, 17.1 versus 9.7 points. The stronger result with 100 tasks therefore reflects fewer disrupted successes rather than more recoveries. This comparison reinforces the need to assess transferable evolution through \mbox{both kinds of outcome change.}

\noindent\textbf{Evolution cost.}\quad
\method\ achieves the highest DeepSWE score, 76.4\%, with the lowest evolution cost among the compared methods, 132.7M tokens (\cref{fig:evolution-efficiency}). The cost difference largely reflects candidate validation. Fresh source executions account for over 99\% of the recorded evolution tokens for Meta-Harness and Self-Harness. \method\ instead replays archived trajectories to examine trigger behavior and reviews guidance for dependence on source-specific assumptions. This design screens for transferable policies with fewer new executions while \mbox{achieving stronger target performance.}

\section{Conclusion}

We presented \method, a framework for evolving harnesses beyond their source benchmark. It combines context-sensitive triggers with diagnostic guidance, using cross-task replay and independent semantic selection to refine interventions. Evolved on an independent task set, \method\ improves strong coding models with both lightweight and mature harnesses and achieves state-of-the-art results on DeepSWE v1.1 and Terminal-Bench 2.1. Gains on compositional OOD tasks further demonstrate transfer to unseen requirement combinations. Ablations and target-task analysis show how selective intervention and evidence-based reassessment support these gains, enabling corrections on new tasks while \mbox{reducing disruption to successful executions.}

\pagebreak
\clearpage

\twocolumn
\section*{Impact Statement}

This work improves coding agents through their execution infrastructure, without updating model weights. More reliable harnesses can reduce repeated debugging and improve automated software development. Because evolved interventions influence code changes and verification, deployment should retain checks appropriate to software security and maintainability alongside functional task evaluation.

\begingroup
\urlstyle{same}
\bibliographystyle{icml2027}
\bibliography{references}
\endgroup

\clearpage
\onecolumn
\appendix

\section{Analysis Details}
\label{app:proofs}

\subsection{Improvement Ceiling and Selectivity}

\begin{figure}[htbp]
\centering
\begin{minipage}[t]{0.48\textwidth}\centering
\begin{tikzpicture}
\begin{axis}[theplot,width=\linewidth,height=180pt,xmin=50,xmax=95,ymin=0,ymax=55,
  xlabel={Baseline success (\%)},ylabel={Maximum gain (points)},
  title={(a) Remaining improvement ceiling},xtick={50,60,70,80,90},ytick={0,10,20,30,40,50}]
\addplot[fill=liftblue!10,draw=none,domain=50:95,samples=80] {100-x} \closedcycle;
\addplot[liftblue,domain=50:95,samples=80] {100-x};
\node[font=\small,anchor=south west,text=liftblue] at (axis cs:73,29) {$1-p_0$};
\end{axis}\end{tikzpicture}
\end{minipage}\hfill
\begin{minipage}[t]{0.48\textwidth}\centering
\begin{tikzpicture}
\begin{axis}[theplot,width=\linewidth,height=180pt,xmin=50,xmax=95,ymin=0,ymax=1.05,
  xlabel={Baseline success (\%)},ylabel={Regression / recovery, $q/r$},
  title={(b) Selectivity required for a gain},xtick={50,60,70,80,90},ytick={0,.25,.5,.75,1}]
\addplot[fill=liftgreen!12,draw=none,domain=50:95,samples=80] {(100-x)/x} \closedcycle;
\addplot[liftgreen,domain=50:95,samples=80] {(100-x)/x};
\node[font=\small,text=liftgreen] at (axis cs:64,.19) {Positive gain};
\node[font=\small,text=liftred] at (axis cs:80,.78) {Net regression};
\node[font=\scriptsize,anchor=south west] at (axis cs:75,.34) {$q/r=(1-p_0)/p_0$};
\end{axis}\end{tikzpicture}
\end{minipage}
\caption{Analytical constraints on harness improvement. (a) The ceiling $1-p_0$ assumes every baseline failure is recovered and no success is lost. (b) For $r>0$, positive gain requires the conditional regression-to-recovery ratio to lie below the boundary in \cref{eq:threshold}. Higher baseline success reduces both the remaining gain and the admissible relative regression. Curves follow the identities above.}
\label{fig:selectivity-boundary}
\end{figure}
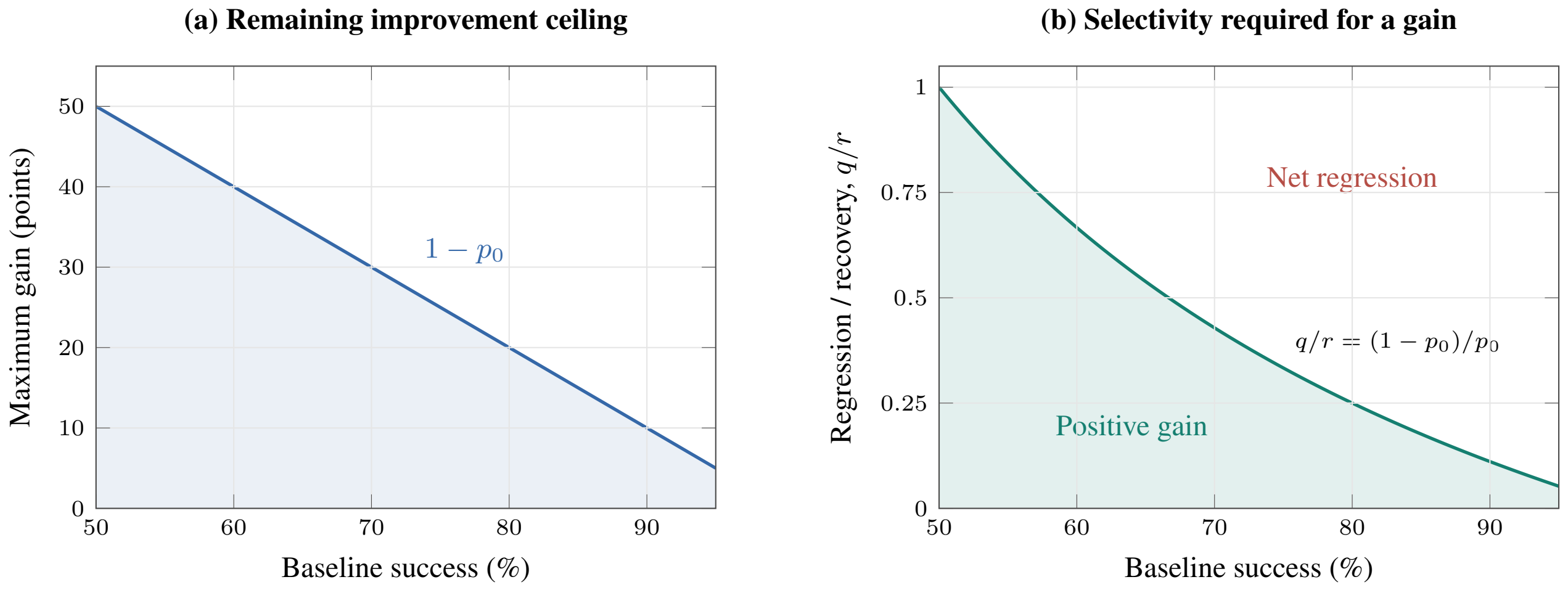

Let $(Y_0,Y_B)$ be paired binary outcomes with base success rate $p_0$, conditional rescue rate $r$, and conditional regression rate $q$. Partitioning the four possible outcome pairs gives
\begin{align}
\E[Y_B-Y_0]
&=\Prob(Y_0=0,Y_B=1)-\Prob(Y_0=1,Y_B=0)\\
&=(1-p_0)r-p_0q.
\end{align}
Because $0\leq r,q\leq1$, the gain is bounded by
\begin{equation}
\Delta=(1-p_0)r-p_0q\leq(1-p_0)r\leq1-p_0.
\label{eq:ceiling}
\end{equation}
For $0<p_0<1$, a positive gain occurs if and only if
\begin{equation}
q<\frac{1-p_0}{p_0}r.
\label{eq:threshold}
\end{equation}
For $q>0$, positive gain equivalently requires $r/q>p_0/(1-p_0)$. Increasing baseline success simultaneously contracts the improvement ceiling and raises the recovery-to-regression ratio required for further gains (\cref{fig:selectivity-boundary}). The achievable rates depend on the model, harness, and task distribution. When $q=0$, any positive recovery rate yields a gain.

At $p_0=0$ or $p_0=1$, one conditional rate is undefined, so we use the unconditional paired probabilities directly. A baseline that never succeeds can only improve or remain unchanged; a baseline that always succeeds can only regress or remain unchanged.

The net gain depends on marginal success rates; its recovery and regression components additionally depend on the pairing. Our empirical analysis pairs executions by task and repeat, as specified in \cref{app:paired-outcomes}.

\subsection{Exposure and Conditional Value}

Consider executions of $H$ and $H+u$ on task $x$ coupled through a common random stream. Assume that their inputs, environment transitions, and effects of resource limits agree until $u$ first acts. Let $A_u$ denote that event. Their histories agree before $A_u$, and their outcomes agree if $A_u$ never occurs. Hence
\begin{equation}
|Y_{H+u}-Y_H|\leq\ind[A_u].
\end{equation}
Conditioning the outcome difference on $A_u$ and its complement yields \cref{eq:scope}. It also gives
\begin{equation}
|\Delta_Q(u\mid H)|\leq\Prob_Q(A_u),
\label{eq:exposure}
\end{equation}
and
\begin{equation}
\Prob_Q(Y_H=1,Y_{H+u}=0)
\leq\Prob_Q(A_u,Y_H=1).
\end{equation}
Under this coupling, a regression requires exposing an otherwise successful execution to the intervention. When $\Prob_Q(A_u)=0$, the expected gain is zero.

The event $A_u$ includes resource effects that can alter outcomes, including trigger computation that exhausts a time limit. Session-wide guidance has exposure probability one. This coupling defines the theoretical decomposition; empirical task-paired statistics use the API executions described in \cref{app:paired-outcomes}.

\subsection{Transfer and Composition}

For $D\in\{P,Q\}$ with positive exposure, write $a_D=\Prob_D(A_u)$ and $v_D=\E_D[Y_{H+u}-Y_H\mid A_u]$. Subtracting the source and target factorizations in \cref{eq:scope} and adding and subtracting $a_Qv_P$ gives
\begin{align}
\Delta_Q(u\mid H)-\Delta_P(u\mid H)
&=a_Qv_Q-a_Pv_P\\
&=(a_Q-a_P)v_P+a_Q(v_Q-v_P).
\end{align}
This is the identity in \cref{eq:transfer-gap}. The decomposition first changes exposure while holding source conditional value fixed, then changes conditional value at target exposure. Both factors can vary across distributions. In particular, $a_Q=a_P$ leaves the gain difference $a_Q(v_Q-v_P)$: the same activation frequency can accompany a lower target value when the behavior being corrected is justified by new requirements. Task-dependent reassessment allows guidance to be interpreted against those requirements. At zero exposure the gain is zero; the conditional value is undefined, and the conditional decomposition is used only where its exposure terms are positive.

For ordered bundle prefixes $B_j=(u_1,\ldots,u_j)$ with $B_0=\varnothing$, the total gain decomposes as
\begin{equation}
\Delta_Q(B_m)=
\sum_{j=1}^m\left(\E_Q[Y_{B_j}]-\E_Q[Y_{B_{j-1}}]\right).
\label{eq:composition}
\end{equation}
Each term is the marginal contribution of an intervention after its predecessors have been applied. Each contribution is conditioned on its predecessors, because composition changes the contexts in which later interventions act. The ordered bundle is therefore the unit of final evaluation.

\section{Method Details}

\subsection{Source Evidence and Candidate Construction}

The author examines complete successful and failed trajectories, including task instructions, model messages, tool calls, and observations. It also receives the runtime interfaces and checking tools. The task repository, reference solution, and hidden grading tests are excluded from the authoring workspace. Each within-task success--failure pair starts from the same base harness, so candidates are developed independently.

DeepSeek Harness replays the motivating pair alongside failed and mixed-outcome executions from other source tasks. About one quarter of the tasks with only failed executions are held out of replay. Cross-task feedback is computed on tasks other than the one that motivated the intervention.

\subsection{Candidate Refinement and Final Selection}
\label{app:modes}

Candidate construction separates authoring review from the final selection in \cref{sec:selection}. The author diagnoses a same-task success--failure contrast and implements an intervention against the base harness. A separate reviewer checks the diagnosis and proposed guidance against both executions. It may edit guidance wording or return changes to the diagnosis or trigger for one author rework. The checker validates the final implementation and records replay behavior. \Cref{tab:review-case} illustrates a refinement that preserves verification while narrowing the repair obligation.

Final selection uses fixed replay criteria followed by an independent semantic assessment. The activation gate in \cref{eq:selection-gate} uses other source-task pairs, excluding the motivating task. The selector bases its verdict on the completed candidate, the original pair, and the harness interface specification. Its decision is independent of cross-task activation statistics and target outcomes, and it accepts or rejects each candidate. After rework, the revised intervention undergoes the same assessment.

Each of the three semantic judgments requires explicit evidence. Diagnosis assessment checks that the cited actions support the claimed defect, that the successful execution does not contradict the account, and that a competing explanation has not been ignored. Source-independence assessment rejects task-specific names, values, or instructions that reveal the successful solution. Guidance assessment permits inspection and verification, while requiring a failed check or a conflicting explicit requirement before a corrective edit. Changes to existing test expectations require an explicit contrary requirement. A candidate is retained only when all three judgments pass; selected members preserve their source order within each interface.

\begin{table}[htbp]
\centering
\caption{Semantic refinement of a completion intervention (\cref{sec:replay}). The entries paraphrase the source example: review preserves verification while changing which failures obligate a repair.}
\label{tab:review-case}
\small\renewcommand{\arraystretch}{1.15}
\begin{tabular}{@{}>{\centering\arraybackslash}p{.16\linewidth}*{2}{>{\centering\arraybackslash}p{.29\linewidth}}@{}}
\toprule
 & Initial guidance & Reviewed guidance \\
\midrule
Verification & Run existing tests. & Run existing tests. \\
Repair scope & Fix any failures. & Repair failures caused by the current change. \\
Unrelated failures & The instruction also obligates repair. & Report the failure without expanding the repair obligation. \\
Decision criterion & A failing test is enough to demand repair. & The failure's relation to the current change determines the response. \\
\bottomrule
\end{tabular}
\end{table}

\subsection{Policies in the Evolved Qwen Code Harness}
\label{app:policy-inventory}

We illustrate construction and selection with Qwen3.8-Flash-Next and Qwen Code. Applying \method\ to 26 source trajectory pairs produced eleven candidates that passed the replay gate. Eight also passed all three independent semantic checks for diagnosis, source independence, and guidance. Of these retained policies, three were revised by the authoring agent, two received wording refinements from the reviewer, and three were accepted as proposed. The policy set and its order were fixed before target evaluation. The evolved harness achieves a mean success rate of 89.9\% on Terminal-Bench 2.1 (\cref{tab:main}). \Cref{tab:policy-inventory} describes each policy by its activation context and diagnostic purpose.

\begin{table}[!htbp]
\centering
\caption{The eight policies retained for Qwen Code. Activation describes when the policy intervenes. Guidance describes what the agent checks using evidence from the current task.}
\label{tab:policy-inventory}
\small
\setlength{\tabcolsep}{5pt}
\renewcommand{\arraystretch}{1.15}
\begin{tabular}{@{}>{\centering\arraybackslash}p{.22\linewidth}>{\centering\arraybackslash}p{.28\linewidth}>{\centering\arraybackslash}p{.45\linewidth}@{}}
\toprule
Policy & Activation & Guidance \\
\midrule
Requirement contract & Once at session start. & Treat observable requirements as the contract. Act on gaps exposed by checks and ground verification in visible evidence. \\
\addlinespace[2pt]
Read truncated output & Once before the next model step after a truncation marker. & Page through the result or narrow the query before deciding an item is absent. \\
\addlinespace[2pt]
Interpret passing baseline checks & Once after a passing check before production code is edited. & Identify existing behavior that should remain and verify the new requirement separately. \\
\addlinespace[2pt]
Reconcile old checks with new requirements & At session start and after the first passing pre-edit check. & Check whether an old assertion encodes a contract superseded by the new requirement. \\
\addlinespace[2pt]
Verify with real project objects & After a self-authored file runs successfully, up to twice. & Exercise actual project objects and helpers, including cases where the relevant member is missing. \\
\addlinespace[2pt]
Audit requirement coverage & After a self-authored check passes following an edit to another file, up to twice. & Exercise valid and alternative forms, missing inputs, and malformed inputs required by the task. \\
\addlinespace[2pt]
Vary the test setup & Once after a passing check involving a self-authored artifact. & Vary assumptions about caller state, completeness, and representation. Change the code if a new case fails. \\
\addlinespace[2pt]
Test suppressed behavior & Once per passing self-authored script with a stub, override, skip, or error suppression. & Check whether the suppressed behavior is required. If so, remove the suppression and rerun verification. \\
\bottomrule
\end{tabular}
\end{table}

\subsection{Selection and Transfer Cases}
\label{app:selection-transfer-cases}

\paragraph{Source selection.}
A candidate for reviewing self-authored checks in Qwen Code diagnosed how a self-written check reinforced an incorrect resampling rule. It passed replay and the semantic checks for diagnosis and source independence. Its guidance, however, treated existing repository behavior as binding whenever the instruction was ambiguous. On the motivating trace, the trigger fired before a requirement check had failed. The selector rejected the candidate because the guidance prescribed an implementation at a point where the available evidence supported further verification. The decision illustrates what semantic selection adds to replay. A well-placed intervention can still prescribe a correction that the task does not support.

\paragraph{Target recovery.}
In Terminal-Bench's \texttt{build-cython-ext} task, the agent must build and install a Python package compatible with NumPy 2.3.0. After adapting its code for renamed graph attributes in a dependency, the evolved DeepSeek Harness prompted the agent to check the closest existing implementation. The agent searched the module and repository for an equivalent normalization helper, found none, and retained the compatibility change. It then reran the repository tests and installed the rebuilt package. The evolved execution passed all eleven benchmark checks. The base execution passed nine, failing the repository and its test-suite checks. The policy transferred a check for an existing implementation to a new compatibility repair.

\subsection{Intervention Scope and Validation}
\label{app:slots}

The runtime exposes five groups of intervention locations. The agent loop accepts guidance at session start or before a model step. Observation and context-management locations add guidance to, or reorganize, successful tool results while retaining the underlying information. The tool interface can deny a pending call once with guidance while permitting an identical retry, or add a tool. The completion interface can request a bounded continuation before termination. These locations determine the information a check can inspect and the actions it can affect.

Validation first checks permitted edits, access to grading information, interface use, and attribution, then compiles and replays the candidate. Replay records actual changes to messages and control flow independently of the candidate's own activation reports. Violations of information preservation, retry permission, or eventual termination reject the candidate. Timing, coverage, and activation in successful executions provide feedback for refinement.

For DeepSeek Harness, accepted policies are composed and their full execution chain is replayed. The runtime fixes the order of interface locations and preserves the selected policy order within each location. Construction and replay use recorded trajectories; fresh source executions subsequently evaluate the composed harness. Applying \method{} to another runtime requires mapping these intervention locations and preserving their execution constraints.

\section{Evaluation Details}
\label{app:eval}

\subsection{Evolution Set Construction}
\label{app:source-protocol}

Our independent evolution set is selected from the 731-task SWE-bench Pro release \citep{swebenchpro2026}. Selection combines the difficulty score and coverage constraints in \cref{eq:source-selection}. Public resolution rates are averaged over available model records, with language offsets $\beta_{\mathrm{JS}}=0$, $\beta_{\mathrm{TS}}=0.02$, $\beta_{\mathrm{Go}}=0.06$, and $\beta_{\mathrm{Python}}=0.10$. These offsets prioritize JavaScript/TypeScript and Go during difficulty-ranked selection.

The initial pass selects tasks by difficulty within each repository: up to two with $r_{\mathrm{pub}}\leq0.15$ and one each in the bands $(0.15,0.40]$ and $(0.40,0.60]$. A global pass fills the set to 100, prioritizing a cap of 28 Python tasks. A final coverage pass raises underrepresented repositories to four tasks with its hardest remaining tasks, replacing tasks from the most represented repository.

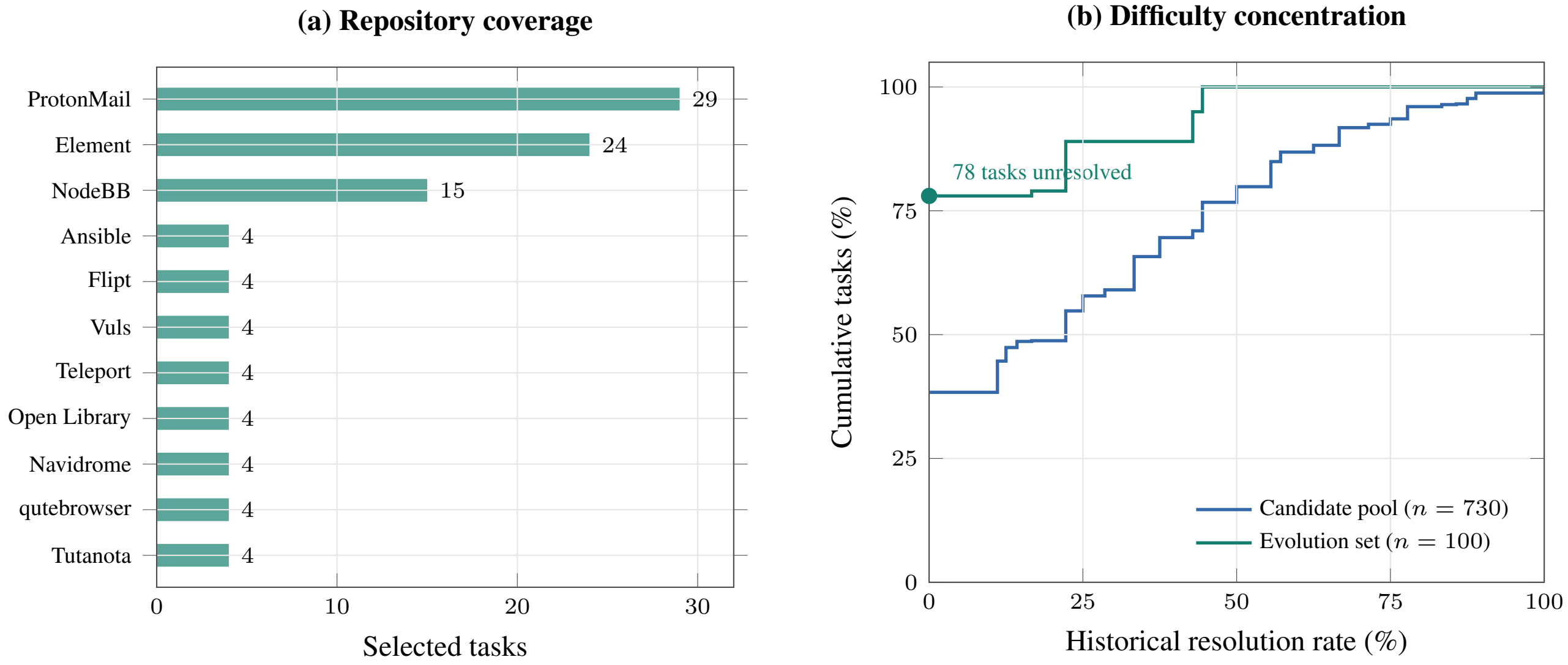
\begin{figure}[htbp]
\centering
\begin{minipage}[t]{0.48\textwidth}\centering
\begin{tikzpicture}
\begin{axis}[theplot,width=.95\linewidth,height=205pt,xmin=0,xmax=32,ymin=-.7,ymax=10.7,
  xbar,bar width=7pt,ytick={0,...,10},yticklabels={ProtonMail,Element,NodeBB,Ansible,Flipt,Vuls,Teleport,Open Library,Navidrome,qutebrowser,Tutanota},
  y dir=reverse,xlabel={Selected tasks},title={(a) Repository coverage},xtick={0,10,20,30},
  nodes near coords,nodes near coords style={font=\scriptsize,anchor=west},enlarge x limits=false]
\addplot[fill=liftgreen!70,draw=none] coordinates {(29,0) (24,1) (15,2) (4,3) (4,4) (4,5) (4,6) (4,7) (4,8) (4,9) (4,10)};
\end{axis}\end{tikzpicture}
\end{minipage}\hfill
\begin{minipage}[t]{0.48\textwidth}\centering
\begin{tikzpicture}
\begin{axis}[theplot,width=\linewidth,height=205pt,xmin=0,xmax=100,ymin=0,ymax=105,
  xlabel={Historical resolution rate (\%)},ylabel={Cumulative tasks (\%)},
  title={(b) Difficulty concentration},xtick={0,25,50,75,100},ytick={0,25,50,75,100},
  legend pos=south east,legend cell align=left]
\addplot[liftblue,const plot] coordinates {(0.0000,38.3562) (11.1100,44.6575) (12.5000,47.3973) (14.2900,48.6301) (16.6700,48.7671) (22.2200,54.7945) (25.0000,57.8082) (28.5700,59.0411) (33.3300,65.7534) (37.5000,69.5890) (42.8600,70.9589) (44.4400,76.7123) (50.0000,79.8630) (55.5600,84.9315) (57.1400,86.8493) (62.5000,88.2192) (66.6700,91.7808) (71.4300,92.4658) (75.0000,93.5616) (77.7800,96.0274) (83.3300,96.4384) (85.7100,96.5753) (87.5000,97.6712) (88.8900,98.7671) (100.0000,100.0000)};
\addlegendentry{Candidate pool ($n=730$)}
\addplot[liftgreen,const plot] coordinates {(0.0000,78.0000) (16.6700,79.0000) (22.2200,89.0000) (42.8600,95.0000) (44.4400,100.0000) (100,100)};
\addlegendentry{Evolution set ($n=100$)}
\addplot[only marks,mark=*,liftgreen,mark size=2pt] coordinates {(0,78)};
\node[anchor=south west,font=\scriptsize,text=liftgreen] at (axis cs:2,79) {78 tasks unresolved};
\end{axis}\end{tikzpicture}
\end{minipage}
\caption{Coverage and difficulty of the independent evolution set. (a) The selected tasks cover all eleven repositories, with at least four tasks per repository. (b) Empirical cumulative distributions of public resolution rates in the 731-task candidate pool and the selected 100 tasks. The selected distribution concentrates on difficult tasks while retaining repository coverage; each task contributes equally. Historical rates use available public model records: 730 candidate tasks and all 100 selected tasks.}
\label{fig:source-selection-profile}
\end{figure}

The resulting coverage and difficulty profile appears in \cref{fig:source-selection-profile}. Each selected task has six to nine public model records, with a mean historical resolution rate of 7.2\%; 78 tasks remain unresolved in these records. The difficulty criterion concentrates trajectories on challenging behavior, while the repository floor spreads that evidence across eleven codebases. The resulting archive provides both failures to learn from and varied contexts in which to assess whether the proposed policies apply.

\subsection{Source Quality Checks}
\label{app:source-quality}

All 100 selected tasks include the statements, environments, and verifier artifacts needed for execution. Each has a nonempty test patch, a reference-test overlay, and a fail-to-pass set; 64 also specify a nonempty pass-to-pass set. Reference tests and solutions remain outside the agent and candidate-author workspaces. The task descriptions used by the runtime originate from the repository change requests in the source release.

We inspect alignment between instructions and tests on one task per source language. The review finds a missing submodule revision and an unrelated passage in an artwork task. The latter is excluded from task-content annotation. Agents receive the original instructions.

\subsection{Source--Target Composition}
\label{app:distribution}

We treat repositories as domains \citep{wilds2021} and each task as one observation. Repository and language labels come from task metadata \citep{deepswe2026}. After standardizing repository identifiers, we find no overlap between the eleven source repositories and 91 DeepSWE repositories. Task identifiers and complete instructions are also disjoint.

Across the two sets, we apply the same rule for mixed JavaScript and TypeScript projects. The resulting language distributions have an empirical total variation distance of 0.366.

\paragraph{Common task-content analysis.}
Task intent and functional requirements provide complementary views of code benchmarks \citep{swebenchpro2026,bigcodebench2025,terminalbench2026}. For \cref{fig:source-distribution}, we apply one codebook to the 100 source, 113 DeepSWE, and 89 Terminal-Bench instructions. Generic benchmark wrappers are excluded. The task-type label identifies the requested final deliverable: repair existing software; extend or restructure it; build a new solution; produce data or analytical results; or configure or restore a system. Adding a previously unsupported mode or format is extension; restoring an already expected behavior is repair. Mixed tasks are assigned by their principal requested deliverable.

Eight non-exclusive labels capture explicit requirements: interface or data-format contracts, state or resource lifecycles (including concurrency), compatibility, system setup and operation, numerical objectives, performance and resource constraints, security and access control, and user-interface behavior.

Model-assisted coding records supporting instruction passages for each positive label. An independent pass labels twelve fixed-seed tasks per task set, and disagreements are resolved against the common definitions. Each task contributes once; requirement labels are non-exclusive, so their proportions can exceed 100\% in total.

\paragraph{Application domain.}
We assign each task one primary application domain based on the purpose of the requested functionality, independently of the type of change or implementation. The six classes are product workflows, programming and development, systems and services, data and content processing, scientific and ML computing, and security and cryptography.

Domain labels use the full instructions and their supporting passages. Independent annotation of twelve tasks per set yields initial agreements of 12/12 for the evolution set, 11/12 for DeepSWE, and 10/12 for Terminal-Bench. We retain alternative labels for two evolution tasks, four DeepSWE tasks, and five Terminal-Bench tasks to assess their effect on the distances.

\begin{table}[t]
\centering
\caption{Application domains across our independent evolution set and the test benchmarks. Each task contributes to one class; entries are percentages of the 100 evolution tasks, 113 DeepSWE tasks, and 89 Terminal-Bench tasks.}
\label{tab:task-context}
\small
\begin{tabular}{cccc}
\toprule
Application domain & Evolution set (TransHarness) & DeepSWE v1.1 & Terminal-Bench 2.1 \\
\midrule
Product workflows & 59.0 & 2.7 & 1.1 \\
Programming and development & 11.0 & 67.3 & 22.5 \\
Systems and services & 15.0 & 10.6 & 12.4 \\
Data and content processing & 3.0 & 13.3 & 19.1 \\
Scientific and ML computing & 1.0 & 4.4 & 34.8 \\
Security and cryptography & 11.0 & 1.8 & 10.1 \\
\bottomrule
\end{tabular}
\end{table}

\paragraph{Distribution distances.}
Using \cref{eq:task-shift}, task-type distances from our independent evolution set are 0.4704 for DeepSWE and 0.9213 for Terminal-Bench; application-domain distances are 0.6996 and 0.6140 (\cref{tab:task-context,fig:source-distribution}). The domain differences persist under arbitrary reassignment of all flagged labels: the corresponding lower bounds remain 0.644 and 0.538. Together with disjoint repositories and differing requirement frequencies, these comparisons characterize the shift under which harness transfer is evaluated.

\paragraph{Within- and cross-benchmark comparison.}
\panelref{fig:source-distribution}{c} compares each evolution set against the same target subset. Inspired by the within-benchmark split in Self-Harness \citep{selfharness2026}, we draw 10,000 uniform random 2:1 partitions per benchmark, yielding evolution and evaluation subsets of 75 and 38 tasks for DeepSWE and 59 and 30 tasks for Terminal-Bench 2.1. For each partition, we measure the distance from the fixed evaluation subset to its complementary within-benchmark evolution subset and to our independent 100-task evolution set.

In DeepSWE, using the independent set increases the mean task-type distance from 0.041 to 0.470 and the mean application-domain distance from 0.128 to 0.704. In Terminal-Bench, the respective distances increase from 0.160 to 0.922 and from 0.184 to 0.639. Matching the evolution-set sizes yields task-type and domain distances of 0.471 and 0.705 on DeepSWE and 0.922 and 0.645 on Terminal-Bench.

\paragraph{Compositional out-of-distribution transfer.}
We call a target task compositionally OOD when its exact nonempty set of requirement labels never occurs in the evolution set. This identifies 16 DeepSWE v1.1 tasks and 49 Terminal-Bench 2.1 tasks. Three Terminal-Bench tasks lack a positive requirement label and contribute to the full-benchmark results.

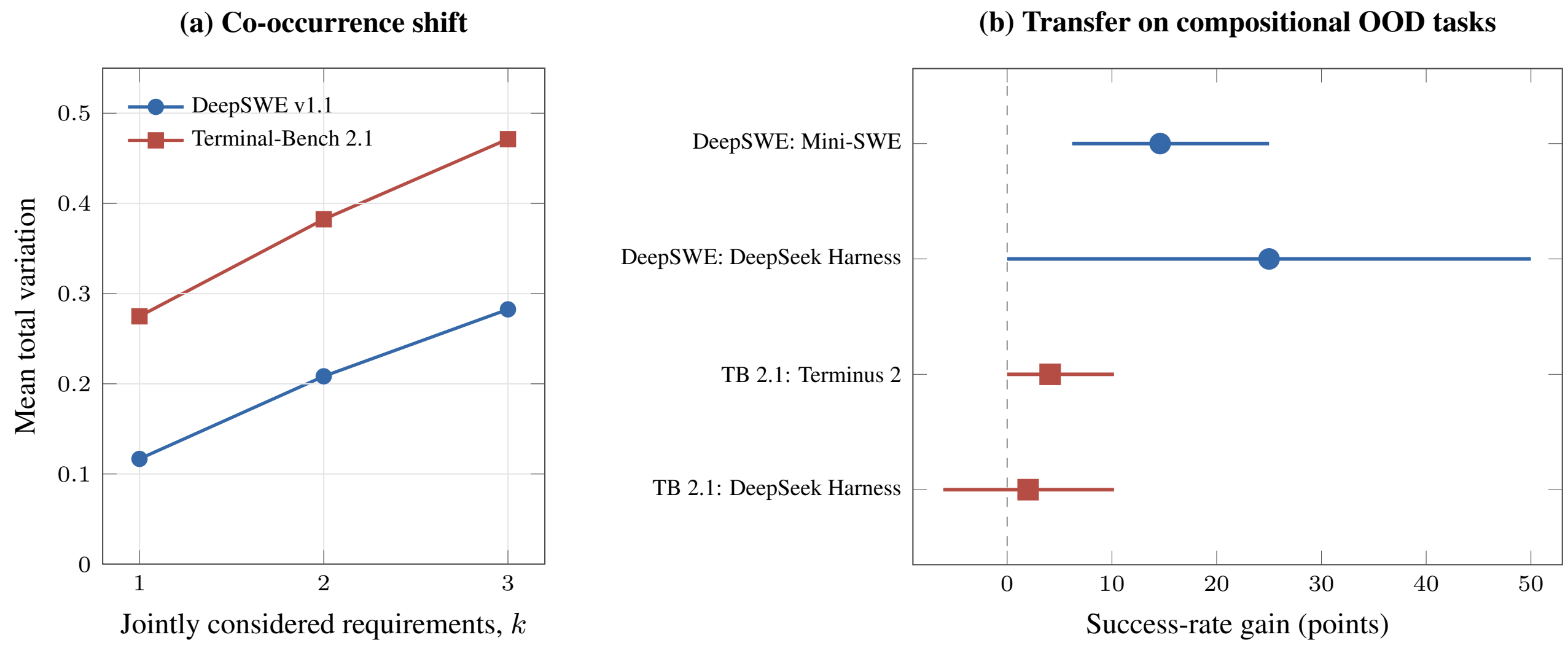
\begin{figure}[!ht]
\centering
\begin{minipage}[t]{0.37\textwidth}\centering
\begin{tikzpicture}
\begin{axis}[theplot,width=\linewidth,height=197.0pt,xmin=.8,xmax=3.2,ymin=0,ymax=.55,
  xtick={1,2,3},ytick={0,.1,.2,.3,.4,.5},xlabel={Jointly considered requirements, $k$},
  ylabel={Mean total variation},title={(a) Co-occurrence shift},legend pos=north west,legend cell align=left]
\addplot[liftblue,mark=*,mark size=2pt] coordinates {(1,0.11672566) (2,0.20830278) (3,0.28251580)};\addlegendentry{DeepSWE v1.1}
\addplot[liftred,mark=square*,mark size=2pt] coordinates {(1,0.27482558) (2,0.38236711) (3,0.47129983)};\addlegendentry{Terminal-Bench 2.1}
\end{axis}\end{tikzpicture}
\end{minipage}\hfill
\begin{minipage}[t]{0.61\textwidth}\centering
\begin{tikzpicture}
\begin{axis}[theplot,width=.82\linewidth,height=197.0pt,xmin=-9,xmax=53,ymin=.35,ymax=4.65,
  xtick={0,10,20,30,40,50},ytick={1,2,3,4},y dir=reverse,
  yticklabels={DeepSWE: Mini-SWE,DeepSWE: DeepSeek Harness,TB 2.1: Terminus 2,TB 2.1: DeepSeek Harness},
  xlabel={Success-rate gain (points)},title={(b) Transfer on compositional OOD tasks},grid=none]
\draw[black!40,dashed] (axis cs:0,.5)--(axis cs:0,4.5);
\draw[liftblue,line width=1.1pt] (axis cs:6.2,1)--(axis cs:25.0,1);
\addplot[only marks,mark=*,mark size=2.8pt,liftblue] coordinates {(14.6,1)};
\draw[liftblue,line width=1.1pt] (axis cs:0,2)--(axis cs:50,2);
\addplot[only marks,mark=*,mark size=2.8pt,liftblue] coordinates {(25,2)};
\draw[liftred,line width=1.1pt] (axis cs:0,3)--(axis cs:10.2,3);
\addplot[only marks,mark=square*,mark size=2.8pt,liftred] coordinates {(4.1,3)};
\draw[liftred,line width=1.1pt] (axis cs:-6.1,4)--(axis cs:10.2,4);
\addplot[only marks,mark=square*,mark size=2.8pt,liftred] coordinates {(2.0,4)};
\end{axis}\end{tikzpicture}
\end{minipage}
\caption{Compositional OOD transfer. (a) Source--target total variation at three sizes of requirement-label subsets. (b) Gains on tasks whose exact requirement combinations are absent from the evolution set. Points and bars show paired mean gains and 95\% task-bootstrap intervals from \cref{tab:requirement-transfer}.}
\label{fig:requirement-transfer}
\end{figure}

We also measure how the frequency of label combinations shifts between source and target. For a subset $A$ of the $m=8$ binary requirements and a pattern $z$, let $p_A(z)$ and $q_A(z)$ be the source and target proportions. We average total variation over subsets of size $k$ \citep{platzer2021}:
\begin{equation}
\overline{\mathrm{TV}}_k=\binom{m}{k}^{-1}\sum_{|A|=k}\frac12\sum_{z\in\{0,1\}^k}|p_A(z)-q_A(z)|.
\end{equation}
We use all 8, 28, and 56 subsets of sizes one, two, and three. \Cref{fig:requirement-transfer} shows larger discrepancies for Terminal-Bench than for DeepSWE at each size.

\Cref{tab:requirement-transfer} reports paired gains for DeepSeek across both benchmarks and both harness types. All four gains are positive on compositional OOD tasks. On DeepSWE, Mini-SWE gains 14.6 points and DeepSeek Harness gains 25.0. On Terminal-Bench, Terminus 2 gains 4.1 points and DeepSeek Harness gains 2.0. The intervals resample tasks within each evaluated subset and preserve the base/evolved pairing. For DeepSeek Harness on DeepSWE, the paired analysis uses the same 70.8\% base score reported in \cref{tab:main}.

\begin{table}[!ht]
\centering
\caption{DeepSeek performance on the full target benchmarks and their compositional OOD subsets. A task is compositional OOD when its exact nonempty requirement-label set is absent from the evolution set. Gains and paired task-bootstrap intervals are in percentage points.}
\label{tab:requirement-transfer}
\small
\renewcommand{\arraystretch}{1.16}
\setlength{\tabcolsep}{7pt}
\newcommand{\gainvalue}[1]{\makebox[2.7em][r]{\ensuremath{#1}}}
\newcommand{\ciendpoint}[1]{\makebox[2.5em][r]{\ensuremath{#1}}}
\newcommand{\pairedinterval}[2]{\ensuremath{[}\ciendpoint{#1}\ensuremath{,}\,\ciendpoint{#2}\ensuremath{]}}
\begin{tabular}{@{}ccccccc@{}}
\toprule
 & & & \multicolumn{2}{c}{Success (\%)} & \multicolumn{2}{c}{Paired gain} \\
\cmidrule(lr){4-5}\cmidrule(l){6-7}
Base harness & Target subset & Tasks & Base & \method & $\Delta$ & 95\% CI \\
\midrule
\rowcolor{black!9}[0pt][0pt]\multicolumn{7}{@{}c@{}}{\textbf{\textit{DeepSWE v1.1}}} \\
\multirow{2}{*}{Mini-SWE} & All tasks & 113 & 73.2 & 76.4 & \gainvalue{+3.2} & \pairedinterval{-2.4}{8.8} \\
 & Compositional OOD & 16 & 72.9 & 87.5 & \gainvalue{+14.6} & \pairedinterval{6.2}{25.0} \\
\addlinespace[3pt]
\multirow{2}{*}{DeepSeek Harness} & All tasks & 113 & 70.8 & 75.2 & \gainvalue{+4.4} & \pairedinterval{-5.3}{14.2} \\
 & Compositional OOD & 16 & 68.8 & 93.8 & \gainvalue{+25.0} & \pairedinterval{0.0}{50.0} \\
\midrule
\rowcolor{black!9}[0pt][0pt]\multicolumn{7}{@{}c@{}}{\textbf{\textit{Terminal-Bench 2.1}}} \\
\multirow{2}{*}{Terminus 2} & All tasks & 89 & 84.3 & 86.5 & \gainvalue{+2.2} & \pairedinterval{-4.5}{9.0} \\
 & Compositional OOD & 49 & 89.8 & 93.9 & \gainvalue{+4.1} & \pairedinterval{0.0}{10.2} \\
\addlinespace[3pt]
\multirow{2}{*}{DeepSeek Harness} & All tasks & 89 & 89.9 & 95.5 & \gainvalue{+5.6} & \pairedinterval{-1.1}{12.4} \\
 & Compositional OOD & 49 & 89.8 & 91.8 & \gainvalue{+2.0} & \pairedinterval{-6.1}{10.2} \\
\bottomrule
\end{tabular}

\end{table}

\subsection{Evaluation Sources and Cost}
\label{app:evolution-tokens}

Public DeepSWE scores use the October 19, 2026 Datacurve snapshot \citep{datacurve2026leaderboard}. Terminal-Bench 2.1 scores use the highest score per listed model in the November 7, 2026 Artificial Analysis snapshot \citep{artificialanalysis2026tb21}. Those evaluations use Terminus 2 with an agent limit of at least 7,200 seconds \citep{artificialanalysis2026methodology}. \Cref{tab:public-configs} lists the reported reasoning settings.

Evolution cost counts input and output tokens for candidate construction and source validation, with cached inputs counted once. We exclude shared archive collection and target evaluation. \Cref{tab:evolution-search-budget} explains the difference in \cref{fig:evolution-efficiency}. Fresh source executions account for over 99\% of the recorded evolution tokens of Meta-Harness and Self-Harness. The measured Mini-SWE construction uses replay and semantic review to screen policies for transfer, with zero new scored source runs.

\begin{table}[!htbp]
\noindent
\begin{minipage}[t]{.445\textwidth}
\vspace{0pt}
\centering
\caption{Reported reasoning settings and agents for the public reference scores in \cref{tab:main}. Settings retain the terminology of their source reports.}
\label{tab:public-configs}
\begingroup
\fontsize{9}{10.8}\selectfont
\renewcommand{\arraystretch}{1.12}
\setlength{\tabcolsep}{3pt}
\begin{tabular}{@{}cccc@{}}
\toprule
Model & \shortstack{DeepSWE\\effort} & \shortstack{TB 2.1\\effort} & \shortstack{TB 2.1\\agent} \\
\midrule
Gemini 3.1 Pro$^{*}$ & high & \emptycell & Terminus 2 \\
Muse Spark 1.1 & xhigh & xhigh & Terminus 2 \\
Grok 4.5 & high & high & Terminus 2 \\
Claude Sonnet 5 & max & max & Terminus 2 \\
Claude Opus 4.8 & max & max & Terminus 2 \\
GPT-5.5 & xhigh & xhigh & Terminus 2 \\
Claude Fable 5 & xhigh & max$^{\dagger}$ & Terminus 2 \\
GPT-5.6 Sol & max & xhigh & Terminus 2 \\
Claude Opus 5 & max & max & Terminus 2 \\
GPT-6 Astra & xhigh & high & Terminus 2 \\
\bottomrule
\end{tabular}
\par\smallskip
{\footnotesize $^{*}$Artificial Analysis reports Gemini 3.1 Pro Preview. $^{\dagger}$The Fable 5 configuration uses Opus 4.8 as a fallback.}
\endgroup

\end{minipage}\hfill
\begin{minipage}[t]{.53\textwidth}
\vspace{0pt}
\centering
\caption{Source-validation budgets for the DeepSWE method comparison. A scored run is one new execution of a source task. Token totals include recorded construction and validation usage. The \method\ cost is measured in a separate construction run using 14 archived trajectory pairs. Self-Harness reports recorded token usage.}
\label{tab:evolution-search-budget}
\small
\setlength{\tabcolsep}{3pt}
\renewcommand{\arraystretch}{1.16}
\newcommand{\budgetcount}[1]{\makebox[2.2em][r]{#1}}
\newcommand{\budgettokens}[1]{\makebox[3.2em][r]{#1}}
\begin{tabular}{@{}>{\centering\arraybackslash}p{.23\linewidth}>{\centering\arraybackslash}p{.34\linewidth}cc@{}}
\toprule
 & \multicolumn{2}{c}{Source validation} & \\
\cmidrule(lr){2-3}
Method & Candidate evaluation & \shortstack{New scored\\runs} & \shortstack{Tokens\\(M)} \\
\midrule
Meta-Harness & One candidate, 100 tasks, two runs per task & \budgetcount{200} & \budgettokens{798.1} \\
Self-Harness & Four updates and one merged harness, 200 runs each & \budgetcount{1,000} & \budgettokens{4,226.3} \\
RRSI & Four candidates, ten tasks each & \budgetcount{40} & \budgettokens{389.0} \\
\addlinespace[3pt]
\textbf{\method\ (Ours)} & Archived-trajectory replay and semantic review & \budgetcount{\textbf{0}} & \budgettokens{\textbf{132.7}} \\
\bottomrule
\end{tabular}
\end{minipage}
\end{table}

\subsection{Compared Methods}
\label{app:compared-methods}

\paragraph{Meta-Harness.} A meta-agent searches over harness code using an archive of prior candidates, outcomes, and execution traces. Source evaluations provide the feedback for ranking harness revisions \citep{metaharness2026}.

\paragraph{Self-Harness.} The agent mines recurring weaknesses from execution traces and proposes targeted changes to its harness. The acceptance procedure checks gains and regressions on development and held-out source tasks \citep{selfharness2026}.

\paragraph{RRSI.} RRSI regularizes both proposal and acceptance. Its search favors sparse, evidence-backed edits and explores underused components. Its selection rule balances measured improvement, stability, and execution cost \citep{rrsi2026}. In our comparison, none of the RRSI candidates met its acceptance rule. We evaluate the candidate with the highest source-task success rate, breaking ties by lower token cost.

\subsection{Scores and Paired Outcomes}
\label{app:paired-outcomes}

For $n$ target tasks and $N$ independent runs of a fixed harness $H$, the reported score is
\begin{equation}
S(H)=\frac{100}{Nn}\sum_{k=1}^{N}\sum_{i=1}^{n}y_{ik}^{H},
\end{equation}
where $y_{ik}^{H}$ is a binary task outcome. We pair base and evolved outcomes by task and evaluation index.

Let $N_{ab}$ count paired executions with base outcome $a$ and evolved outcome $b$. We report recovered and regressed percentages as $100N_{01}/(Nn)$ and $100N_{10}/(Nn)$. Their difference equals the net score gain. The corresponding conditional rates are $r=N_{01}/(N_{01}+N_{00})$ and $q=N_{10}/(N_{10}+N_{11})$, giving recovered and regressed percentages of $100(1-p_0)r$ and $100p_0q$. All executions enter these counts, including those without an activated check. A paired task bootstrap resamples whole tasks, preserving the conditions and all repetitions within each task.

We apply a common scoring rule across methods. Verifier or environment repairs preserve the submitted implementation and test assertions. Agent errors, exhausted budgets, and incomplete test reports count as failures. Executions interrupted solely by infrastructure faults are rerun.

\begin{table}[!htbp]
\noindent
\begin{minipage}[t]{.52\textwidth}
\vspace{0pt}
\centering
\caption{Repeated DeepSWE v1.1 evaluation with DeepSeek V4.1 Flash (1017) and Mini-SWE. Each run contains all 113 tasks. Success and gain are percentages and percentage points, respectively.}
\label{tab:repeat-results}
\small\setlength{\tabcolsep}{3.5pt}\renewcommand{\arraystretch}{1.15}
\begin{tabular}{@{}ccccc@{}}
\toprule
Run & Base solved & \shortstack{\method\\solved} & \shortstack{\method\\score} & Gain \\
\midrule
1 & 80 & 87 & 77.0 & +6.2 \\
2 & 83 & 87 & 77.0 & +3.5 \\
3 & 85 & 85 & 75.2 & 0.0 \\
\midrule
Mean & & & 76.4 & +3.2 \\
\bottomrule
\end{tabular}
\end{minipage}\hfill
\begin{minipage}[t]{.445\textwidth}
\vspace{0pt}
\Cref{tab:repeat-results} reports paired outcomes for the three independent evaluation runs. The gain ranges from 0.0 to 6.2 percentage points across the three runs. Overall, \method\ solves 259 of 339 executions, compared with 248 for the base harness, with 44 recoveries and 33 regressions. Resampling the 113 tasks with all paired repetitions in 50,000 bootstrap replicates gives 95\% intervals of $[69.9,82.6]$ for success and $[-2.4,8.8]$ for the gain. The across-run standard deviation is 1.0 percentage point.
\end{minipage}
\end{table}

\subsection{Intervention Design Comparisons}
\label{app:ablations}

The execution variants in \cref{tab:mechanism} separate the roles of context selection and guidance formulation within the evolved harness. The variant without learned triggers delivers each intervention at its first eligible interface event. The variant without reassessment preserves the learned trigger and diagnostic subject and replaces diagnostic guidance with a prescribed correction. Task requirements and evidence remain part of that guidance. Interfaces, ordering, activation caps, and global policies are shared, so these comparisons examine how the same diagnostic subject enters execution and shapes the agent's response.

The unreviewed construction instead examines the harness obtained from fourteen author-produced interventions after interface validation. The main configuration combines seven reviewed interventions. This comparison evaluates the resulting content and composition together, complementing the execution variants with evidence about construction feedback. The source example in \cref{sec:replay} illustrates the underlying revision: verification remains required, while repair is scoped to failures caused by the current change.

\Needspace{6\baselineskip}\subsection{Source-Budget Settings}
\label{app:efficiency}

\Cref{tab:efficiency} compares harnesses constructed from nested source subsets with three archived executions per task. The 25-, 50-, and 75-task conditions use patch extraction, fixed activation thresholds, and semantic screening; the 100-task condition is the main seven-intervention harness. All conditions use the complete DeepSWE target and the same paired-outcome definitions.

\section{Extended Related Work}
\label{app:related}

\paragraph{From feedback to reusable knowledge.}
Feedback can improve an agent's current answer or change what it carries into subsequent tasks. Self-Refine iteratively critiques and revises outputs with the same model \citep{selfrefine2023}, while CRITIC grounds correction in external tool feedback \citep{critic2024}. Reflexion retains verbal feedback across attempts \citep{reflexion2023}; ExpeL extracts reusable insights from task experience \citep{expel2024}. These approaches establish complementary roles for feedback: identifying a local mistake and retaining knowledge that can guide later decisions. \method\ builds on this distinction at the harness level. Its diagnostic guidance identifies an issue to reassess, and its executable trigger determines when that issue becomes relevant in a new execution. The resulting policy transfers a procedure for examining evidence, allowing the correction itself to depend on the target task.

\paragraph{The object of optimization.}
Harnesses organize coding agents' reasoning, tools, and verification \citep{react2023,codeact2024}. 
Prompt and program optimization use evaluation feedback to improve instructions or compositions of language-model calls \citep{opro2024,dspy2024,gepa2026}, while automated agent design searches broader executable structures \citep{adas2025,aflow2025}. Harness evolution makes the execution system itself editable, including its tools, control flow, and verification behavior \citep{sica2025,dgm2026,metaharness2026}. Recent work structures this search through module-level attribution, constrained component edits, or knowledge that guides task-local evolution \citep{modularrsi2026,harnesscompass2026,meta22026}. Within this design space, \method\ treats a trigger and its diagnostic guidance as a coupled policy. This representation makes two questions explicit: which observed execution contexts warrant intervention, and what evidence should the agent examine there? Cross-task replay evaluates the first question against both successful and failed source executions. Semantic selection examines the second by checking the diagnosis and whether the guidance remains meaningful without the source task's particular solution.

\paragraph{Selection evidence and transfer.}
Candidate selection determines which source-derived changes survive into deployment. Regression checks and regularized acceptance can discourage harmful updates \citep{selfharness2026,rrsi2026}, while executable verification can test proposed improvements during evolution \citep{verse2026}. Their effect on new tasks also depends on how search feedback relates to the final evaluation distribution. Separating development feedback from held-out evaluation is therefore central to interpreting agent improvement \citep{rethinkingharnessevaluation2026,harnessdev2026,ding2026ade}. Our independent evolution set extends this separation across benchmarks, with task-content analysis characterizing the resulting distribution shift. The empirical argument combines that setting with target success, paired recovery and regression, and intervention ablations: distribution analysis establishes where transfer is tested, while target outcomes and controlled comparisons establish its value and explain the role of the evolved policies.

\end{document}